\documentclass[10pt,journal,compsoc]{IEEEtran}
\usepackage[table]{xcolor}
\ifCLASSOPTIONcompsoc
  \usepackage[nocompress]{cite}
\else
  \usepackage{cite}
\fi
\ifCLASSINFOpdf
\else
\fi
\usepackage{amsmath,amsfonts,bm}

\def\eqref#1{equation~\ref{#1}}

\def\1{\bm{1}}

\def\vd{{\bm{d}}}

\def\vh{{\bm{h}}}

\def\vm{{\bm{m}}}

\def\vx{{\bm{x}}}

\DeclareMathAlphabet{\mathsfit}{\encodingdefault}{\sfdefault}{m}{sl}
\SetMathAlphabet{\mathsfit}{bold}{\encodingdefault}{\sfdefault}{bx}{n}

\def\vx{{\bm{x}}}
\def\vx{{\bm{x}}}

\newcommand{\Dtrain}[1]{D^{tr}_{#1}}

\newcommand{\ACC}{ACC($\uparrow$) }
\newcommand{\FM}{FM($\downarrow$) }
\newcommand{\LA}{LA($\uparrow$) }
\newcommand{\BWT}{BWT($\uparrow$) }

\newcommand{\Splus}{$\mc S^+$}
\newcommand{\Sminus}{$\mc S^-$}
\newcommand{\Sin}{$\mc S^{\mathrm{in}}$}
\newcommand{\Sout}{$\mc S^{\mathrm{out}}$}
\newcommand{\Spl}{$\mc S^{\mathrm{pl}}$}

\newcommand{\mc}{\mathcal}

\newcommand{\KLD}[2]{D_{\mathrm{KL}} \left( \left. \left. #1 \right|\right| #2 \right) }

\usepackage{booktabs}
\usepackage{multirow}
\usepackage{graphicx,subcaption,ragged2e}
\usepackage{cellspace}
\usepackage[utf8]{inputenc} 
\usepackage[T1]{fontenc}    
\usepackage[colorlinks=true,linkcolor=blue,allcolors=blue]{hyperref}
\usepackage{url}            
\usepackage{booktabs}       
\usepackage{amsfonts}       
\usepackage{nicefrac}       
\usepackage{microtype}      
\usepackage{xcolor}         
\usepackage{graphicx}
\usepackage{amsmath,bm}
\usepackage{amssymb, bbold}
\usepackage{multirow}
\usepackage{subcaption}
\usepackage{mathtools}
\usepackage{array}
\usepackage{xr}
\def\1{\bm{1}}
\def\vx{{\bm{x}}}

\def\vM{{\bm{M}}}
\def\vZ{{\bm{Z}}}

\usepackage[ruled,vlined,linesnumbered]{algorithm2e}
\SetKwInput{kwInit}{Init}
\SetKwInput{kwFunc}{function}
\SetKwInput{kwEnd}{end function}
\SetKwInput{kwRequire}{Require}
\SetKwComment{Comment}{$\triangleright$\ }{}

\SetCommentSty{mycommfont}
\newcommand{\nonl}{\renewcommand{\nl}{\let\nl\oldnl}}
\newcommand{\SubItem}[1]{
	{\setlength\itemindent{15pt} \item[-] #1}
}
\newcommand{\blue}[1]{#1}
\newcommand{\red}[1]{#1}
\begin{document}
%
\title{Continual Learning, Fast and Slow}
%
%
%
%

\author{Quang~Pham, Chenghao~Liu, Steven C. H. Hoi, \textit{Fellow, IEEE}
\IEEEcompsocitemizethanks{
\IEEEcompsocthanksitem Corresponding author: Quang Pham is with Institute for Infocomm Research (I$^2$R), A$^*$Star. \protect\\
E-mail: \texttt{hqpham.2017@phdcs.smu.edu.sg} 
\IEEEcompsocthanksitem Chenghao Liu is with Salesforce Research Asia. \protect\\
E-mail: \texttt{chenghao.liu@salesforce.com}
\IEEEcompsocthanksitem Steven C. H. Hoi is with Singapore Management University and Salesforce Research Asia.
\protect \\ 
E-mail: \texttt{chhoi@smu.edu.sg} }
\thanks{Manuscript received MM DD, YYYY; revised MM DD, YYYY.}}

%
%

\markboth{Journal of \LaTeX\ Class Files,~Vol.~14, No.~8, August~2015}%
{Shell \MakeLowercase{\textit{et al.}}: Bare Demo of IEEEtran.cls for Computer Society Journals}
%



\IEEEtitleabstractindextext{%
}
\IEEEtitleabstractindextext{
\begin{abstract}
According to the Complementary Learning Systems (CLS) theory~\cite{mcclelland1995there} in neuroscience, humans do effective \emph{continual learning} through two complementary systems: a fast learning system centered on the hippocampus for rapid learning of the specifics, individual experiences; and a slow learning system located in the neocortex for the gradual acquisition of structured knowledge about the environment. Motivated by this theory, we propose \emph{DualNets} (for Dual Networks), a general continual learning framework comprising a fast learning system for supervised learning of pattern-separated representation from specific tasks and a slow learning system for representation learning of task-agnostic general representation via Self-Supervised Learning (SSL). DualNets can seamlessly incorporate both representation types into a holistic framework to facilitate better continual learning in deep neural networks. Via extensive experiments, we demonstrate the promising results of DualNets on a wide range of continual learning protocols, ranging from the standard offline, task-aware setting to the challenging online, task-free scenario. 
Notably, on the CTrL~\cite{veniat2020efficient} benchmark that has unrelated tasks with vastly different visual images, DualNets can achieve competitive performance with existing state-of-the-art dynamic architecture strategies~\cite{ostapenko2021continual}. Furthermore, we conduct comprehensive ablation studies to validate DualNets efficacy, robustness, and scalability. Code will be made available at \url{https://github.com/phquang/DualNet}.

\end{abstract}
}
\maketitle

\IEEEdisplaynontitleabstractindextext

%
\IEEEpeerreviewmaketitle

\begin{figure}[t]
	\centering
	\includegraphics[width=0.45\textwidth]{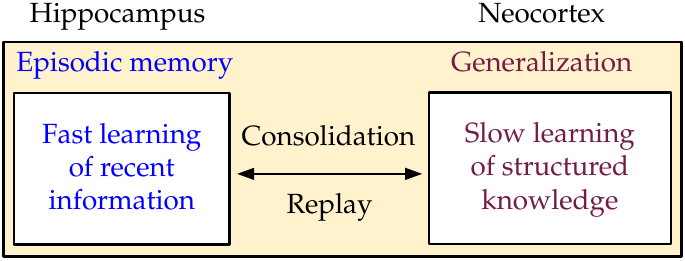}
	\caption{An illustration of the fast and slow learning according to the CLS theory. Figure is inspired from~\cite{parisi2019continual}.} \label{fig:cls}
	\vspace*{-0.1in}
\end{figure}

\IEEEraisesectionheading{\section{Introduction}\label{sec:intro}}
Humans have the remarkable ability to learn and accumulate knowledge over their lifetime to perform different cognitive tasks. Interestingly, such a capability is attributed to the complex interactions among different interconnected brain regions~\cite{douglas1995recurrent}. 
One prominent model is the \emph{Complementary Learning Systems (CLS) theory}~\cite{mcclelland1995there,kumaran2016learning} which suggests \blue{that much of the learning and remembering capabilities are attributed to the interactions between the systems located at the hippocampus and neocrotex regions.}\footnote{\blue{The brain functionalities involve many components~\cite{o2014complementary} and understanding its mechanism is still an open research topic. For simplicity and brevity, we use hippocampus and neocortex to refer to the ``hippocampus-dependent" and ``neocortex-dependent" systems.}} Particularly, the hippocampus focuses on fast learning of pattern-separated representation of specific experiences~\cite{barry2019remote,tonegawa2018role}. Via the memory consolidation process, the hippocampus's memories are \blue{consolidated} to the neocortex over time to form a more general representation that supports long-term retention and generalization to new experiences.
The two fast and slow learning systems constantly interact to facilitate fast learning and long-term remembering, which is illustrated in Fig.~\ref{fig:cls}.
On the other hand, although deep neural networks have achieved impressive results in many applications~\cite{lecun2015deep}, they often require having access to a large amount of data while performing poorly when learning on data streams~\cite{sahoo2018online,aljundi2019online}. Moreover, when the stream consists of data from several distributions, deep neural networks \emph{catastrophically forget} past knowledge, which further hinders the overall performance~\cite{french1999catastrophic,kirkpatrick2017overcoming,lopez2017gradient}. Therefore, the main focus of this study is exploring how the CLS theory can motivate a general continual learning framework with a better trade-off between alleviating catastrophic forgetting and facilitating knowledge transfer.

In literature, several continual learning strategies are inspired from the CLS theory, from using the episodic memory~\cite{lopez2017gradient} to improving the representation~\cite{javed2019meta,rao2019continual}. 
\blue{However, most of them use supervised learning to model both the hippocampus and neocortex functionalities, which lack a separate slow, general representation learning component.} During continual learning, the representation obtained by repeatedly performing supervised learning on a small amount of memory data can be prone to overfitting and may not generalize well across tasks. On the other hand. recent studies in continual learning show that unsupervised representation~\cite{gepperth2016bio,parisi2018lifelong} is often more resisting to forgetting compared to the supervised representation, which yields little improvements~\cite{javed2018revisiting}.
This result motivates us to conceptualize a novel fast-and-slow learning framework for continual learning, which comprises a two separate learning systems. The fast learner focuses on supervised learning while the slow learner focuses on accumulating better representations. As a result, the fast learner can take advantage of the slow representation to learn new tasks more efficiently, while retaining the old tasks' knowledge.

\begin{figure*}[t]
	\begin{center}
		\centerline{\includegraphics[width=1.\textwidth]{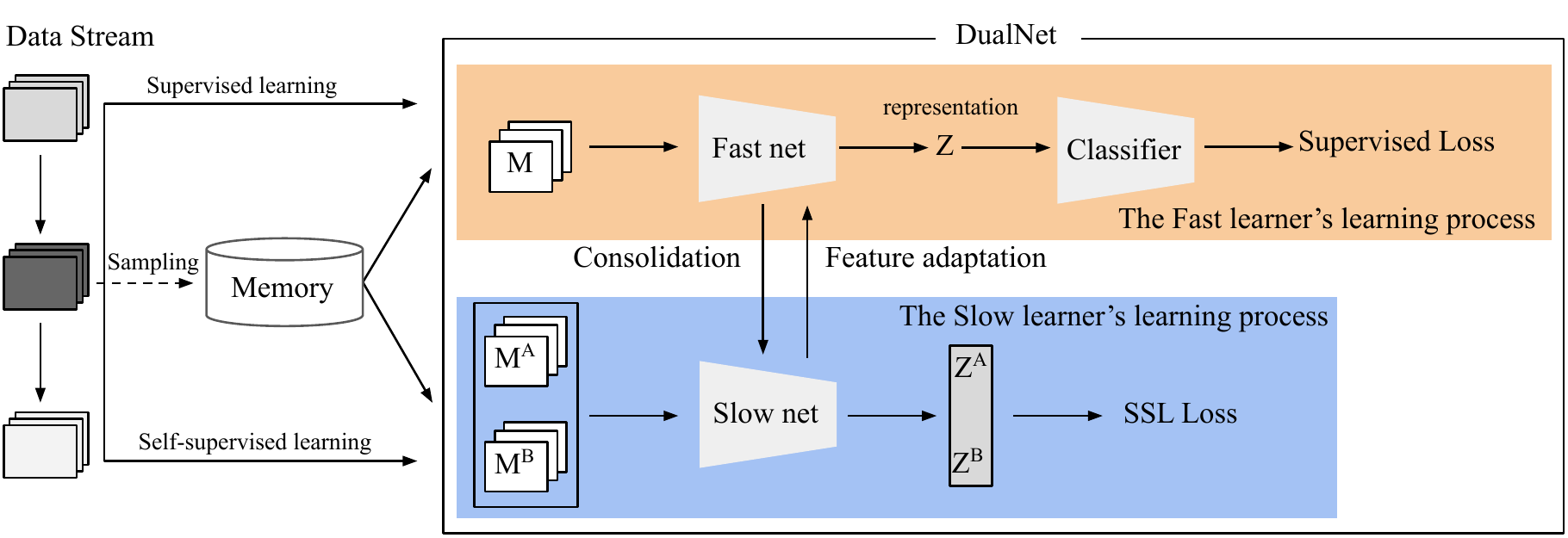}}
		
		\caption{Overview of the DualNet architecture, which consists of (i) a slow learner (blue) that learns representation by optimizing an SSL loss using samples from the memory, and (ii) a fast learner (orange) that adapts the slow net's representation for quick knowledge acquisition of labeled data. Both learners can be trained synchronously. $\vM$ denotes a randomly sampled mini-batch and $\vM^A$, $\vM^B$ denote two views of $\vM$ obtained by applying two different data transformations. Best viewed in colors.}
		\label{fig:DualNet}
	\end{center}
	\vspace*{-0.2in}
\end{figure*}

From the fast-and-slow learning framework, we propose {\it DualNets} (for Dual Networks), a novel and practical continual learning paradigm consisting of two separate learning systems.
Particularly, DualNets consist of two \emph{complementary} and \emph{parallel} training processes. First, the representation learning phase involves only the \emph{slow network}, which continuously optimizes a Self-Supervised Learning (SSL) loss to model the generic, task-agnostic features~\cite{gepperth2016bio,parisi2018lifelong}. Separating the slow learning phase allows DualNets continuously improve representation even when there are no labeled samples, which is ubiquitous in real-world deployment scenarios where labeled data are delayed~\cite{diethe2019continual} or even limited, which we will demonstrate in Sec.~\ref{sec:semi}.
Secondly and simultaneously, the supervised learning phase involves both learners. In this phase, the goal is to train the fast learner, a more lightweight model that can do supervised learning more efficiently on data streams. We also propose a simple feature adaptation mechanism so that the fast learner can incorporate the slow representations into its predictions to ensure good results.
Fig.~\ref{fig:DualNet} depicts an overview of our DualNets.

Lastly, by design, the original DualNet~\cite{pham2021dualnet} utilizes all slow features to learn the current sample. We note that this strategy may hinder the performance when the continuum (data stream) contains unrelated tasks. In such scenarios, there might be negative transfer among tasks, resulting in a performance drop when using all slow features. In continual learning literature, this challenge is commonly addressed by a dynamic architecture design, which uses different subnetworks for the unrelated tasks~\cite{veniat2020efficient} and can perform well where tasks are not related. However, despite strong results, such strategies are often expensive to train, incurs additional complexities overhead, and often perform poorly in the online learning scenario~\cite{chaudhry2019agem}.
\blue{Since DualNet's slow learning is generic, we propose to enrich DualNet with a more robust fast feature to tackle the complex transfer continual learning scenarios. To this end, we propose DualNet++, which equips DualNet with a simple yet elegant regularization strategy to alleviate the negative knowledge transfer in continual learning. Particularly, DualNet++ inserts a dropout layer between the fast and slow learners' interaction, which prevents its fast learner from co-adapting the slow features. As a result, DualNet++ is robust to the negative knowledge transfer under the presence of unrelated or interference tasks \textbf{without the need for modularized representations}. As we will empirically verify in Sec.~\ref{sec:batch-cl}, DualNet++ achieves promising performance on the CTrL benchmark~\cite{veniat2020efficient}, specifically designed to test the model's ability to transfer knowledge in different complex scenarios.}


In summary, our work makes the following contributions:
\begin{enumerate}
    \item We propose DualNet, a novel and generalized continual learning framework comprising two key components of fast and slow learning systems, which is motivated by the CLS theory.
    \item We develop to practical algorithms of DualNet and DualNet++, which implements the fast and slow learning approaches for continual learning. Notably, DualNet++ is also robust to the negative knowledge transfer.
    \item We conduct extensive experiments to demonstrate DualNet's competitive performance compared to state-of-the-art (SOTA) methods. We also provide comprehensive studies of DualNet's efficacy, robustness to the slow learner's objectives, and scalability to the computational resources. 
\end{enumerate}

\section{Related Work}\label{sec:related}
\subsection{Continual learning}
\blue{Existing continual learning methods~\cite{parisi2019continual,delange2021continual} can be broadly categorized into {\it two} groups.} First, {\it dynamic architecture} methods aims at having a separate subnetwork for each task, thus eliminating catastrophic forgetting to a great extend. The task-specific network can be identified simply allocating new parameters~\cite{rusu2016progressive,yoon2018lifelong,learn2grow}, finding a configuration of existing blocks or activations in the backbone~\cite{fernando2017pathnet,hat-cl}, or generating the whole network conditioning on the task identifier~\cite{von2019continual}. While achieving strong performance, they are often expensive to train and do not work well on the online continual learning setting~\cite{chaudhry2019agem} because of the lack of knowledge transfer mechanism across tasks. In the second category of {\it fixed architecture} methods, learning is regularized by employing a memory to store information of previous tasks. In {\it regularization-based} methods, the memory stores the previous parameters and their importance estimations\cite{kirkpatrick2017overcoming,zenke2017continual,aljundi2017memory,ritter2018online}, which regulates training of newer tasks to avoid changing crucial parameters of older tasks. Recent works have demonstrated that the {\it experience replay} (ER) principle~\cite{lin1992self} is an effective approach and its variants~\cite{lopez2017gradient,chaudhry2019agem,riemer2018learning,liu2020mnemonics,van2018generative,buzzega2020dark} have achieved promising results in different domains, from vision~\cite{chaudhry2019tiny}, language~\cite{de2019episodic,sun2019lamol}, to reinforcement learning~\cite{rolnick2019experience}. 
Recent approaches augment the ER strategies with a calibration component also showed encouraging results. Particularly, they introduce a separate controller to calibrate the backbone network to balance between alleviating forgetting and facilitating knowledge transfer~\cite{pham2021contextual,yin2021mitigating}. It is worth noting that such strategies requires the task identifiers and are only applicable to the task-aware scenarios.

\blue{\textbf{Neuroscience inspired continual learning} \; Neuroscience and the CLS theory have inspired several recent continual learning strategies~\cite{pham2021contextual,aranilearning,hayes2020remind,kemker2018fearnet}. Such studies try to emulate the hippocampus and neocortex by introducing additional learning components~\cite{arani2021learning,pham2021contextual} or memory units~\cite{hayes2020remind,kemker2018fearnet}. The most related work to ours is CLSER~\cite{arani2021learning}, which approaches fast-and-slow learning as updating the learners at different frequencies, i.e. the fast/slow learner is optimized more/less frequently. Nevertheless, CLSER still only learns the supervised learning patterns. Instead of learning at different frequencies, our work approaches the CLS theory from the representation learning perspective by introducing a slow learning phase to the slow net, which was not explored in previous studies. We will empirically compare DualNets with CLSER in Sec.~\ref{sec:batch-cl}.}

\subsection{Representation Learning for Continual Learning}
Representation learning has been an important research field in machine learning and deep learning~\cite{erhan2010does,bengio2013representation}. Recent works demonstrated that a general representation could transfer well to many downstream tasks~\cite{oord2018representation}, or generalize well under limited training samples~\cite{finn2017model}. For continual learning, extensive efforts have been devoted to learning a generic representation that can alleviate forgetting while facilitating knowledge transfer. The representation can be learned either by supervised learning~\cite{rebuffi2017icarl}, unsupervised learning~\cite{gepperth2016bio,parisi2018lifelong,rao2019continual}, or meta (pre-)training~\cite{javed2019meta,he2019task}. While unsupervised and meta training have shown promising results on simple datasets such as MNIST and Omniglot, they lack the scalability to real-world benchmarks. 
\blue{Concurrent with our work, several studies have explored incorporating SSL into continual learning and achieved promising results~\cite{fini2022self,cha2021co2l,bhat2022task}. Different from such studies which use one network for both representations, our DualNets decouple the slow learning into the slow learner, which does not require any pre-training steps and is capable of training synchronously. We will explore the benefits of having two learners in Sec.~\ref{sec:abla-dual}.}

\subsection{Feature Adaptation}
Feature adaptation allows the feature to quickly change and adapt~\cite{perez2018film}. Existing continual learning methods have explored the use of task identifiers~\cite{pham2021contextual,von2019continual,hat-cl} or the memory data~\cite{he2019task} as a context to facilitate learning. While the task identifier is powerful since it provides additional information regarding the task of interest, it is limited only to the task-aware setting or require inferring the underlying task, which can be challenging in practice. On the other hand, sample-based context conditioning is useful in incorporating information of similar samples to the current query and has found success beyond continual learning~\cite{finn2017model,requeima2019fast,dumoulin2018feature-wise,rebuffi2017learning}.
Several continual learning strategies have adopted this approach by implementing the meta-learning gradient updating rules~\cite{he2019task,caccia2020online,pham2020bilevel} and have shown promising results.
However, we argue that naively adopting this approach is not practical for real-world continual learning because the model always performs the full forward/backward during inference, which suffers from the high computational costs. 
Moreover, the predictions are not deterministic because of the dependency on the data chosen for finetuning during inference. For DualNets, feature adaptation plays an important role in the interaction between the fast and slow learners. We address the limitations of existing techniques by developing a novel mechanism that allows the fast learner to efficiently utilize the slow representation without additional information about the task identifiers. Notably, our approach is built upon the feature-wise transformation~\cite{perez2018film}, which does not require backpropagation during testing.

\section{Method}\label{sec:method}
\subsection{Setting and Notations}
We consider the continual learning setting~\cite{lopez2017gradient,chaudhry2019agem} over a continuum $\mc D= \{\vx_i, t_i, y_i\}_i$, where each instance is a labeled sample $\{\vx_i,y_i\}$ with an \emph{optional} task identifier $t_i$. Each labeled sample is drawn from an underlying distribution $P^t(\bm X, \bm Y)$ that represents a task and can suddenly change to $P^{t+1}$, indicating a task switch. In the {\it task-aware setting}, the task identifier $t$ is provided and only the corresponding task's classifier is selected for inference~\cite{lopez2017gradient}. When the task identifier is not provided (during both training and evaluation), the model has a shared classifier for all classes observed so far, which follows the {\it task-free setting}~\cite{chaudhry2018riemannian,aljundi2019task}. We consider both scenarios in our experiments. Note that there is a hybrid setting with task identifiers provided during training but not during evaluation~\cite{ostapenko2021continual}, which we will consider as \emph{task-aware} in this work.

A common continual learning component is the episodic memory $\mc M$ to store a subset of observed data and interleave them when learning the current samples~\cite{lopez2017gradient,chaudhry2019tiny}.   
From $\mc M$, we use $\vM$ to denote a randomly sampled mini-batch, and $\vM^A$, $\vM^B$ to denote two views of $\vM$ obtained by applying two different data transformations.
We also denote $\bm \phi$ as the parameter of the slow network that learns general representation from the input data and $\bm \theta$ as the parameter of the fast network that learns the transformation coefficients. 

\subsection{The DualNets Paradigm}
DualNet learns generic representations to support better generalization capabilities across both old and new tasks in continual learning. 
The model consists two main learning modules (Figure~~\ref{fig:DualNet}): (i) the slow learner is responsible for learning a general representation; and (ii) the fast learner learns with labeled data from the continuum to quickly capture the new information and then consolidate the knowledge to the slow learner. 

DualNets' learning can be broken down into two {\it synchronous} phases. First, the self-supervised learning phase in which the slow learner optimizes an SSL objective on incoming samples and samples from the episodic memory.
Second, the supervised learning phase, which involves the fast learner using the representation from the slow learner and adapting it for supervised learning. The incurred loss will be backpropagated through both learners for supervised knowledge consolidation.  Additionally, the fast learner's adaptation is per-sample-based and does not require additional information such as the task identifiers.  
While the SSL can make the slow learner's representation generic, backprogating the supervised learning loss end-to-end ensures that the slow learner can learn representations that are useful for the supervised learning.
Lastly, DualNet uses the same episodic memory's budget as other methods to store the samples and their labels, but the slow learner only requires the samples while the fast learner uses both samples and their labels.

\begin{figure*}[t]
	\begin{center}
		\centerline{\includegraphics[width=0.95\textwidth]{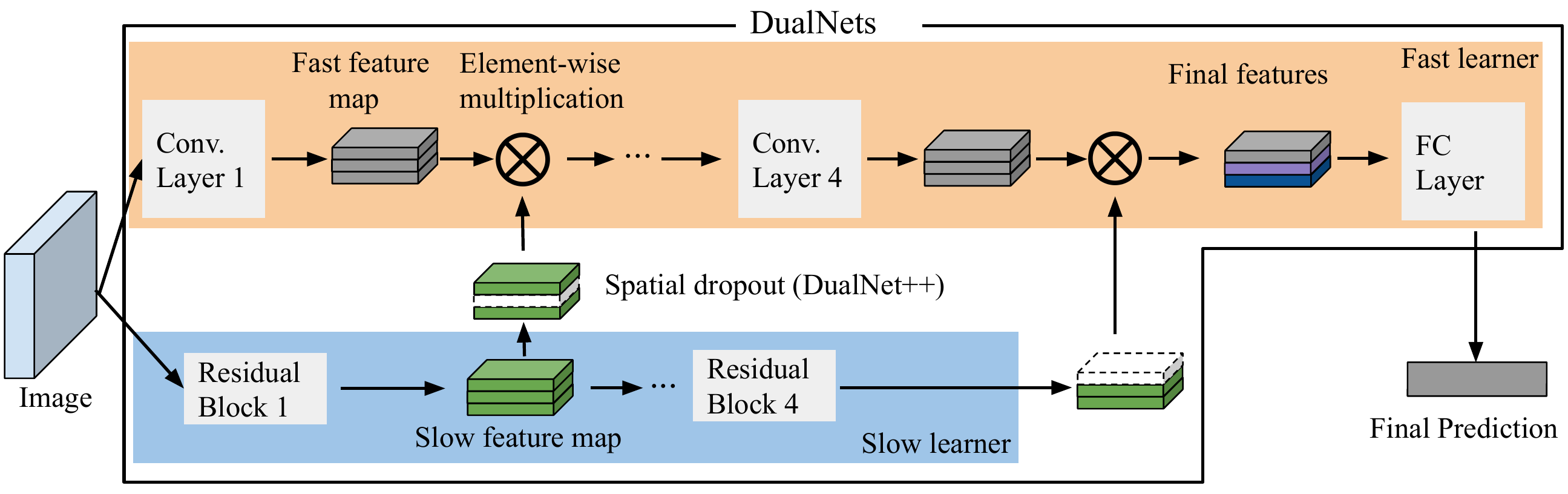}}
		
		\caption{An illustration of DualNets forward calculation during the supervised learning or inference phase on a standard ResNet~\cite{he2016deep} backbone. Given an input image, the slow learner (blue) first performs the forward pass to obtain the feature maps. Then, the fast learner (orange) perform its forward pass using both the slow and fast features. Best viewed in colors.}
		\label{fig:interaction}
	\end{center}
\end{figure*}

\subsubsection{The Slow Learner}\label{sec:slow-learner}
The slow learner is a standard backbone network $\bm \phi$ trained to optimize an SSL loss, denoted by $\mc L_{SSL}$. As a result, any SSL objectives can be applied in this step. However, to minimize the additional computational resources while ensuring a general representation, we only consider the SSL loss that (i) does not require additional memory unit (such as the negative queue in MoCo~\cite{he2020momentum}), (ii) does not always maintain an additional copy of the network (such as BYOL~\cite{grill2020bootstrap}), and (iii) does not use handcrafted pretext losses (such as RotNet~\cite{erhan2010does} or JiGEN~\cite{carlucci2019domain}). Therefore, we consider contrastive SSL losses~\cite{oord2018representation} and implement DualNets using Barlow Twins~\cite{zbontar2021barlow}, a common SSL method that achieved promising results with minimal computational overheads. Formally, Barlow Twins requires two views $\vM^A$ and $\vM^B$ by applying two different data transformations to a batch of images $\vM$. By default, $\vM$ contains incoming samples from the environment and samples in the episodic memory to maximize the number of samples for SSL.
The augmented data are then passed to the slow net $\bm \phi$ to obtain two representations $\vZ^A$ and $\vZ^B$. The Barlow Twins loss is defined as:
\begin{equation}\label{eqn:bt}
    \mc L_{\mc B \mc T} \triangleq \sum_{i}(1-\mc C_{ii})^2 + \lambda_{\mc B \mc T} \sum_i \sum_{j \neq i} \mc C_{ij}^2,
\end{equation}
where $\lambda_{\mc B \mc T}$ is a trade-off factor, and $\mc C$ is the cross-correlation matrix between $\vZ^A$ and $\vZ^B$:
\begin{equation}\label{eqn:cross-correlation}
    \mc C_{ij} \triangleq \frac{\sum_b z^A_{b,i} z^B_{b,j}}{\sqrt{\sum_B (z^A_{b,i})^2} \sqrt{\sum_B (z^B_{b,j})^2} }
\end{equation}
with $b$ denotes the mini-batch index and $i,j$ are the vector dimension indices. 
Intuitively, by optimizing the cross-correlation matrix to be identity, Barlow Twins enforces the network to learn essential information that is invariant to the distortions (unit elements on the diagonal) while eliminating the redundancy information in the data (zero element elsewhere).
In our implementation, we follow the standard practice in SSL to employ a projector on top of the slow network's last layer to obtain the representations $\vZ^A, \vZ^B$. For supervised learning with the fast network, which will be described in Sec.~\ref{sec:fast-learner}, we use the slow network's last layer as the representation $\vZ$. 

\textbf{Optimization in Online Continual Learning }
In most SSL training, the LARS optimizer~\cite{you2017large} is employed for distributed training across many devices, which takes advantage of a large amount of unlabeled data. However, in \textbf{online continual learning}, the episodic memory only stores a small number of samples, which are always changing because of the memory updating mechanism. As a result, the data distribution in the episodic memory always drifts after each iteration, and the SSL loss in DualNet presents different challenges compared to the traditional SSL optimization. Particularly, although the SSL objective in continual learning can be easily optimized using one device, we need to quickly capture the knowledge of the currently stored samples before the newer ones replace them. In this work, we propose to optimize the slow learner using the {\it Look-ahead} optimizer~\cite{zhang2019lookahead}, which performs the following updates:
\begin{align}
    \tilde{\bm \phi}_k \gets& \tilde{\bm \phi}_{k-1} - \epsilon \nabla_{\tilde{\bm \phi}_{k-1}} \mc L_{\mc B \mc T}, \; \mathrm{ with } \; \tilde{\bm \phi}_0 \gets \bm \phi \;\mathrm{ and }\; k = 1,\ldots,K \\
    \bm \phi \gets& \bm \phi + \beta (\tilde{\bm \phi}_K - \bm \phi), 
\end{align}
where $\beta$ is the Look-ahead's learning rate and $\epsilon$ is the Look-ahead's SGD learning rate. As a special case of $K=1$, the optimization reduces to the traditional optimization of $\mc L_{\mc B \mc T}$ using SGD. By performing $K>1$ updates using a standard SGD optimizer, the look-ahead weight $\tilde{\bm \phi}_K$ is used to perform a momentum update for the original slow learner $\bm \phi$. As a result, the slow learner optimization can explore regions that are undiscovered by the traditional optimizer and enjoys faster training convergence~\cite{zhang2019lookahead}. Note that SSL focuses on minimizing the training loss rather than generalizing this loss to unseen samples, and the learned representation requires to be adapted to perform well on a downstream task. Therefore, such properties make the Look-ahead optimizer a more suitable choice over the standard SGD to train the slow learner. For the batch continual learning setting~\cite{kirkpatrick2017overcoming}, because the model can learn the current task for many epochs, it is sufficient to train the slow learner with the standard SGD.

Lastly, we emphasize that although we choose to use Barlow Twins as the SSL objective, DualNets are compatible with any existing methods in the literature, which we will explore empirically in Sec.~\ref{sec:exp-slow-obj}. Moreover, we can always train the slow learner in the background by optimizing Equation~\ref{eqn:bt} synchronously with the continual learning of the fast learner, which we will detail in the following section.

\subsubsection{The Fast Learner}\label{sec:fast-learner}
Given a labeled sample $\{\vx, y\}$, the fast learner's goal is utilizing the slow learner's representation to learn this sample via an adaptation mechanism. 
We propose a general context-free adaptation mechanism by extending and improving the channel-wise transformation~\cite{perez2018film,pham2021contextual} to the general continual learning setting. Particularly, such strategies relies on low dimensional context vectors, such as task identifiers, to learn a channel-wise transformation coefficients. In the task-free setting, the model needs to learn such information from the raw, high dimensional images. To compensate for the increased learning complexity, we propose to implement the fast net as a smaller neural network instead of a simple linear layer~\cite{pham2021contextual}. Moreover, the transformation is pixel-wise instead of channel-wise to allow for a more fine-grained usage of the slow feature given the current image.

Formally, let $\{\vh_i\}_{i=1}^L$ be the feature maps from the slow learner's layers on the image $\vx$, e.g. $\vh_1, \vh_2, \vh_3, \vh_4$ are outputs from four residual blocks in ResNets~\cite{he2016deep}, our goal is to obtain the adapted feature $h'_L$ conditioned on the image $\vx$. Therefore, we design the fast learner as a simple CNN with $L$ layers, and the adapted feature $\vh'_L$ is obtained as
\begin{align}\label{eqn:fastslow}
    \vm_{l} =& g_{\bm \theta,l}(\vh'_{l-1}),  \; \mathrm{with} \; \vh'_0 = \vx \; \mathrm{and} \; l=1,\ldots,L \notag \\
    \vh'_{l} =& \vh_{l} \odot \vm_l, \quad \forall l=1,\ldots,L,
\end{align}
where $\odot$ denotes the element-wise multiplication, $g_{\bm \theta, l}$ denotes the $l$-th layer's output from the fast network $\bm \theta$ and has the same dimension as the corresponding slow feature $\vh_l$. 
The fast net final layer's transformed feature $\vh'_L$ will be fed into a classifier for prediction.

Thanks to the simplicity of the transformation, the fast learner is light-weight but still can take advantage of the slow learner's rich representation for better supervised learning. Meanwhile, the slow learner is mostly trained by the SSL loss to obtain a generic representation that is resistant to catastrophic forgetting. Figure~\ref{fig:interaction} illustrates the fast and slow learners' interaction during the supervised learning or inference phase.

\textbf{The Fast Learner's Objective } 
To further facilitate the fast learner's knowledge acquisition during supervised learning, we also mix the current sample with previous data in the episodic memory, which is a form of experience replay (ER). Particularly, given the incoming labeled sample $\{\vx,y\}$ and a mini-batch of memory data $\bm M$ belonging to a past task $k$, we consider the ER with a soft label loss~\cite{van2018generative} for the supervised learning phase as:
\begin{align}\label{eqn:supervised}
    \mc L_{tr} =& \mathrm{CE}(\pi(\mathrm{DualNet}(\vx),y) +  \frac{1}{|\bm M|} \sum_{i=1}^{|\bm M|} \mathrm{CE}(\pi(\hat{y}_i),y_i) + \notag \\
    +&\lambda_{tr} \KLD {\pi \left( \frac{\hat{y}_i}{\tau} \right) }{\pi \left( \frac{\hat{y}_k}{\tau} \right) },
\end{align}
where $\mathrm{CE}$ is the cross-entropy loss , $D_{\mathrm{KL}}$ is the KL-divergence, $\hat{y}$ is the DualNet's prediction, $\hat{y}_k$ is snapshot of the model’s logits (the fast learner's prediction) of the corresponding sample at the end of task $k$, $\pi(\cdot)$ is the softmax function with temperature $\tau$, and $\lambda_{tr}$ is the trade-off factor between the soft and hard labels in the training loss. Similar to~\cite{pham2021contextual,buzzega2020dark}, Equation~\ref{eqn:supervised} requires minimal additional memory to store the soft label $\hat{y}$ in conjunction with the image $\vx$ and the hard label $y$. 

\subsection{DualNet++}
\blue{This section details DualNet++, an improved version of DualNet to tackle the challenges of complex transfer scenarios in continual learning~\cite{veniat2020efficient}.} In many real-world applications, data in continual learning may contain vastly different visual features, which could even be adversarial to one another~\cite{wang2021afec}. Common approaches to tackle such challenges~\cite{veniat2020efficient,ostapenko2021continual} are mainly based on the dynamic architectures, which compartmentalize knowledge into different modules. Then, the model only uses the relevant subnetwork to learn a current task, which only transfers useful knowledge while alleviating the negative effects of unrelated features. \blue{In contrast, since DualNet already learned the slow, generic feature, we propose to improve the fast features to be more robust to tackle the complex transfer challenges without needing modularized representations.}

To this end, we analyze the DualNet's drawback in achieving a good transfer when learning from complex continual learning streams. We argue that since the slow features in DualNets are obtained from all tasks' data, the original DualNet will learn the fast features dependent on the slow features. While this is helpful in the controlled environments with no negative transfers, it will hinder the performance when there are tasks unrelated to one another. Thus, we propose DualNet++ that alleviates the co-adaptation between the fast and slow features, allowing it to learn more robust features that are useful for the current task. DualNet++ introduces a simple dropout layer~\cite{hinton2012improving} between the fast and slow learners' interactions.
As a result, under the presence of negative transfer, the fast learner will not become dependent on the slow feature~\cite{hinton2012improving,lecun2015deep} and can focus on learning features useful for the current inputs.

To implement DualNet++, we insert a spatial dropout layer~\cite{tompson2015efficient} between the fast and slow learner interaction. Formally, DualNet++ replace the interaction in Eq.~\ref{eqn:fastslow} as: 
\begin{align}
    \vm_{l} =& g_{\bm \theta,l}(\vh'_{l-1}),  \; \mathrm{with} \; \vh'_0 = \vx \; \mathrm{and} \; l=1,\ldots,L \notag \\
    \vh'_{l} =& \bm D_l \odot \vh_l \odot \vm_l, \quad \forall l=1,\ldots,L,
\end{align}
where $\bm D_l \in \mathbb{R}^{n\times w \times h}$ is a spatial dropout mask obtained as
\begin{align}
    d_{l,i} \sim& \mathrm{Bernoulli}(p), \forall i=1,\ldots,n  \label{eqn:ber}\\
    \bm D_l \gets& \mathrm{Repeat}(\vd_l, n\times w\times h), \vd_l = \{d_{l,1},\ldots,d_{l,n}\} \label{eqn:drop}.
\end{align}
In Eq.~\ref{eqn:ber}, $\mathrm{Bernoulli(p)}$ denotes a sample randomly drawn from a Bernoulli distribution with probability $p$, and $\mathrm{Repeat}$ in Eq.~\ref{eqn:drop} denotes reshaping a vector to a particular dimensions by repeating its values along the required axes.
Specifically, the dropout mask $\bm D$ is obtained by performing $n$ independent dropout trial on a feature map of size $n \times w \times h$, and each trial will zero an entire channel. We also note that it is possible to apply the traditional dropout~\cite{hinton2012improving} on the pixels independently, or inserting dropout in the backbone networks. However, preliminary results of such strategies are not promising due to the incompatibility between dropout and batch normalization~\cite{ioffe2015batch, li2019understanding}. Therefore, we decided to not explore these configurations further.
In contrast, the spatial dropout is more suitable for convolutional neural networks, and is only inserted in the fast and slow networks' interaction, not between the hidden layers. Lastly, a recent work~\cite{mirzadeh2020dropout} also show promising results of applying dropout in continual learning. However, their studies only focus on the simple feed-forward architectures and left the convolution networks unexplored.

\section{Experiments}\label{sec:exp}
We compare DualNets against competitive continual learning approaches in both the online~\cite{lopez2017gradient} and offline~\cite{kirkpatrick2017overcoming,veniat2020efficient} scenarios. Our goal of the experiments is to investigate the following hypotheses: (i) DualNets can work well across different continual learning scenarios; (ii) DualNets are robust to the choice of the SSL loss; (iii) DualNets are scalable with the number of SSL training iterations; (iv) DualNet++ can efficiently learn under the presence of unrelated tasks and distribution shifts. In all experiments, DualNet++'s dropout ratio is set as $p=0.1$ for the online setting, and $p=0.2$ for the batch setting, unless otherwise stated.

\subsection{Online Continual Learning Experiments}\label{sec:benchmark}
We first consider the \emph{Online Continual Learning setting}~\cite{lopez2017gradient} where both the tasks and samples within each task arrive sequentially. This setting presents a unique challenge where the catastrophic forgetting and facilitating knowledge transfer problems are entangled~\cite{buzzega2020dark}. Thus, successful online continual learning solutions must achieve a good trade-off of these conflicting objectives. 
\subsubsection{Setup}
\textbf{Benchmarks } We consider the ``Split" continual learning benchmarks constructed from the miniImageNet~\cite{vinyals2016matching}, CORE50 dataset~\cite{core50}, \blue{and the ImageNet} datasets~\cite{russakovsky2015imagenet} with 17, 10, \blue{and 10 continual learning tasks, respectively}. Each task is created by randomly sampling without replacement five classes from the original dataset. We also consider both the task-aware and task-free protocols. \blue{We reserved three tasks from the miniImagenet dataset to cross-validate the hyper-parameters of all methods.} \red{The hyper-parameter configuration is then fixed during continual learning.} We also found that these hyper-parameters could transfer well to the remaining benchmarks.
For the task-aware (TA) protocol, the task identifier is available, and only the corresponding classifier is selected for training and evaluation. In contrast, the task-identifiers are not given in the task-free (TF) protocol, and the models have to predict all classes observed so far. 

\textbf{Evaluation Metrics } We run the experiments five times and report the averaged accuracy of all tasks/classes at the end of training~\cite{lopez2017gradient} (\ACC), the forgetting measure~\cite{chaudhry2018riemannian} (\FM) and backward transfer~\cite{lopez2017gradient} (\BWT), and the learning accuracy~\cite{riemer2018learning} (\LA). Let $a_{i,j}$ be the model's accuracy evaluated on the testing data of task $j$ after it is trained on task $i$. The above metrics are defined as:

\begin{itemize}
	\item{\bf Average Accuracy \ACC (higher is better):}
	\begin{equation}
		\mathrm{ACC}(\uparrow) = \frac{1}{T} \sum_{i = 1}^T a_{T,i}.    
	\end{equation}
	\item{\bf Forgetting Measure \FM (lower is better):}
	\begin{equation}\label{eq:fm}
		\mathrm{FM}(\downarrow) = \frac{1}{T-1} \sum_{j=1}^{T-1} \max_{l \in \{1,\dots T-1\}}a_{l,j} - a_{T,j}.
	\end{equation}
	\item{\bf Backward Transfer \BWT (higher is better):}
	\begin{equation}\label{eq:bwt}
		\mathrm{BWT}(\downarrow) = \frac{1}{T-1} \sum_{j=1}^{T-1} a_{T,j} - a_{j,j},
	\end{equation}
	\item{\bf Learning Accuracy \LA (higher is better):} 
	\begin{equation}
		\mathrm{LA}(\uparrow) = \frac{1}{T} \sum_{i=1}^T a_{i,i}.
	\end{equation}
\end{itemize}

The above metrics provide a comprehensive evaluation of online continual learning methods. Particularly, \ACC reports the overall performance at the end of learning, and is usually used to compared among methods. \FM and \BWT report the average performance drop of each task at the end of learning and indicate the model's ability to address catastrophic forgetting. Lastly, \LA measures the model's ability to learn new tasks indicating its ability to facilitate forward knowledge transfer. Notably, the \LA metric we considered here is an unnormalized variant of the Intransigence Measure~\cite{chaudhry2018riemannian}.

\textbf{Baselines } We compare DualNet and DualNet++ against a suite of state-of-the-art continual learning methods. First, we consider ER~\cite{chaudhry2019tiny}, a simple experience replay method that works consistently well across benchmarks. Then we include DER++~\cite{buzzega2020dark}, an ER variant that augments ER with a $\ell_2$ loss on the soft labels. We also compare with CTN~\cite{pham2021contextual} a recent state-of-the-art method on the online task-aware setting.
For all methods, the hyper-parameters are selected by performing grid-search on the cross-validation tasks.

\textbf{Architecture } We use a full ResNet18~\cite{he2016deep} as the backbone for all methods.
In addition, we construct the DualNet's fast learner as follows: the fast learner has the same number of convolutional layers as the number of residual blocks in the slow learners. A residual block and its corresponding fast learner's layer will have the same output dimensions. With this configuration, the fast learner's architecture is uniquely determined by the slow learner's network and only increased the number of parameters by about 20\%.
Lastly, all networks in this experiment are trained from scratch.

\textbf{Training } In the supervised learning phase, all methods are optimized by the (SGD) optimizer \textbf{over one epoch} with mini-batch size 10 on the Split miniImageNet and \blue{batch 32 for the CORE50 Split Imagenet benchmarks respectively.} In the representation learning phase, we use the Look-ahead optimizer~\cite{zhang2019lookahead} to train the DualNets' slow learner as described in Sec.~\ref{sec:slow-learner}. We employ an episodic memory with \textit{50 samples per task} and the Ring-buffer management strategy~\cite{lopez2017gradient} in the task-aware setting. 
In the task-free setting, the memory is implemented as a reservoir buffer~\cite{vitter1985random} with \text{100 samples per class}. We simulate the synchronous training property in DualNet by training the slow learner with $n$ iterations using the episodic memory data before observing a mini-batch of labeled data. Except for ImageNet, each experiments are repeated for five times.

\textbf{Data pre-processing } DualNet's slow learner follows the data transformations used in BarlowTwins~\cite{zbontar2021barlow}. For the supervised learning phase, we consider two options. First, the standard data pre-processing of no data augmentation during both training and evaluation, which is commonly implemented in existing studies~\cite{lopez2017gradient,chaudhry2019tiny}. Second, we also train the baselines with data augmentation for a fair comparison. However, we observe the data transformation in~\cite{zbontar2021barlow} is too aggressive; therefore, we only implement the random cropping and flipping for the supervised training phase of these baselines.
In all scenarios, the inference phase does not any use data augmentations.

\subsubsection{Results of Online Continual Learning Benchmarks}\label{sec:main-results}
\begin{table*}[t]
\centering
\caption{Evaluation metrics on the Split miniImageNet and CORE50 benchmarks. All methods use an episodic memory of 50 samples per task in the TA setting, and 100 samples per class in the TF setting. The ``Aug" suffix denotes using data augmentation. We highlight the methods with best mean metrics in bold, and underline the second best methods}\label{tab:main}
\begin{tabular}{@{}lcccccc@{}}
\toprule
\multirow{2}{*}{Method} & \multicolumn{3}{c}{Split miniImageNet-TA}       & \multicolumn{3}{c}{Split miniImageNet-TF}       \\ \cmidrule(l){2-7} 
                        &\ACC         &\FM          &\LA          &\ACC         &\FM          &\LA          \\ \midrule
ER~\cite{chaudhry2019tiny}                      & 58.24{\tiny $\pm$0.78} & 9.22{\tiny $\pm$0.78} & 65.36{\tiny $\pm$0.71} & 25.12{\tiny $\pm$0.99} & 28.56{\tiny $\pm$1.10} & 49.04{\tiny $\pm$1.56} \\
ER-Aug                  & 59.80{\tiny $\pm$1.51} & 4.68{\tiny $\pm$1.21}  & 58.94{\tiny $\pm$0.69} & 27.94{\tiny $\pm$2.44}   & 29.36{\tiny $\pm$3.23}        & 54.02{\tiny $\pm$1.02}            \\
DER++~\cite{buzzega2020dark}                   & 62.32{\tiny $\pm$0.78} & 7.00{\tiny $\pm$0.81}  & 67.30{\tiny $\pm$0.57} & 27.16{\tiny $\pm$1.99} & 34.56{\tiny $\pm$2.48} & 59.54{\tiny $\pm$1.53}            \\
DER++-Aug               & 63.48{\tiny $\pm$0.98} & 4.01{\tiny $\pm$1.21}  & 62.17{\tiny $\pm$0.52}  & 28.26{\tiny $\pm$1.81}  & 36.70{\tiny $\pm$1.85}  & 62.70{\tiny $\pm$0.41}          \\
CTN~\cite{pham2021contextual}                     & 65.82{\tiny $\pm$0.59} & {\bf 3.02{\tiny $\pm$1.13}}  & 67.43{\tiny $\pm$1.37} & N/A          & N/A          & N/A          \\
CTN-Aug                 & 68.04{\tiny $\pm$1.23} & 3.94{\tiny $\pm$0.98}  & 69.84{\tiny $\pm$0.78} & N/A          & N/A          & N/A          \\ \midrule
DualNet~\cite{pham2021dualnet}                 & \underline{73.20{\tiny $\pm$0.68}} & \underline{3.86{\tiny $\pm$1.01}}  & \underline{74.12{\tiny $\pm$0.12}} &
\underline{36.86{\tiny $\pm$1.36}} & \underline{28.63{\tiny $\pm$2.26}} & \underline{63.46{\tiny $\pm$1.97}} \\ 
DualNet++ & {\bf 74.24{\tiny $\pm$0.95}} & {\bf 2.83{\tiny $\pm$0.71}}  & {\bf 74.11{\tiny $\pm$0.30}} &
{\bf 37.56{\tiny $\pm$1.12}} & {\bf 27.13{\tiny $\pm$1.16}} & {\bf 63.96{\tiny $\pm$1.02}} \\ \midrule
\multirow{2}{*}{Method} & \multicolumn{3}{c}{CORE50-TA}              & \multicolumn{3}{c}{CORE50-TF}              \\ \cmidrule(l){2-7} 
                        &\ACC         &\FM          &\LA          &\ACC         &\FM          &\LA          \\ \midrule
ER~\cite{chaudhry2019tiny}                      & 41.72{\tiny $\pm$1.30} & 9.10{\tiny $\pm$0.80}  & 48.18{\tiny $\pm$0.81} & 21.80{\tiny $\pm$0.70} & 14.42{\tiny $\pm$1.10} & 33.94{\tiny $\pm$1.49} \\
ER-Aug                  & 44.16{\tiny $\pm$2.05} & 5.72{\tiny $\pm$0.02}  & 47.83{\tiny $\pm$1.61} & 25.34{\tiny $\pm$0.74}            & 15.28{\tiny $\pm$0.63}            & 37.94{\tiny $\pm$0.91}            \\
DER~\cite{chaudhry2019tiny}                     & 46.62{\tiny $\pm$0.46} & 4.66{\tiny $\pm$0.46}  & 48.32{\tiny $\pm$0.69} & 22.84{\tiny $\pm$0.84} & 13.10{\tiny $\pm$0.40}  & 34.50{\tiny $\pm$0.81} \\
DER++-Aug               & 45.12{\tiny $\pm$0.68} & 5.02{\tiny $\pm$0.98}  & 47.67{\tiny $\pm$0.08} & 28.10{\tiny $\pm$0.80} & 10.43{\tiny $\pm$2.10} & 36.16{\tiny $\pm$0.19}            \\
CTN~\cite{pham2021contextual}                     & 54.17{\tiny $\pm$0.85} & 5.50{\tiny $\pm$1.10}  & 55.32{\tiny $\pm$0.34} & N/A          & N/A          & N/A          \\
CTN-Aug                 & 53.40{\tiny $\pm$1.37} & 6.18{\tiny $\pm$1.61}  & 55.40{\tiny $\pm$1.47} & N/A          & N/A          & N/A          \\ \midrule
DualNet~\cite{pham2021dualnet}                 & \underline{57.64{\tiny $\pm$1.36}} & \underline{4.43{\tiny $\pm$0.82}}  & \underline{58.86{\tiny $\pm$0.66}} & \underline{38.76{\tiny $\pm$1.52}} & \underline{8.06{\tiny $\pm$0.43}}  & \textbf{40.00{\tiny $\pm$1.67}} \\ 
DualNet++                 & \textbf{59.07{\tiny $\pm$1.30}} & \textbf{2.86{\tiny $\pm$0.92}} & \textbf{59.23{\tiny $\pm$1.03}} & {\bf 39.42{\tiny $\pm$1.80}} & {\bf 7.08{\tiny $\pm$2.25}}  & \underline{39.52{\tiny $\pm$1.09}} \\
\bottomrule
\end{tabular}
\end{table*}

\begin{table}[t]
\centering
\caption{Evaluation metrics on the task-free Split ImageNet benchmark. Best result is in bold, second best is underlined} \label{tab:imnet}
{\begin{tabular}{lccc}
\toprule
\multirow{2}{*}{Method} & \multicolumn{3}{c}{ImageNet} \\ \cmidrule{2-4}
 & ACC & FM & LA \\ \midrule
ER & 12.30 & 31.30 & 40.50 \\
DER++ & 13.70 & 30.90 & 40.90 \\
DualNet & \underline{14.87} & \underline{30.07} & \textbf{42.60} \\
DualNet++ & \textbf{14.90} & \textbf{29.60} & \underline{41.70} \\ \bottomrule
\end{tabular}}
\end{table}

We report the results of the online continual learning experiments in Tab.~\ref{tab:main} and Tab.~\ref{tab:imnet}.
Our DualNet's slow learner optimizes the Barlow Twins objective for $n=3$ iterations between every incoming mini-batch of labeled data. We will report the impact of the SSL iteration in Sec.~\ref{sec:abla-iter}.

Generally, data augmentation creates more samples to train the models and provides improvements in all cases.
Consistent with previous studies, we observe that DER++ performs slightly better than ER thanks to its soft-label loss. Similarly, CTN can perform better than both ER and DER++ because of its ability to model task-specific features. 
Overall, our DualNets consistently outperform other baselines by a large margin, even with the data augmentation propagated to their training. 
Specifically, DualNets are more resistant to catastrophic forgetting (lower FM) while greatly facilitating knowledge transfer (higher LA), which results in better overall performance, indicated by higher ACC. We also observe that DualNet++ performs marginally better than DualNet in all cases, suggesting the benefits of the spatial dropout regularization.
Additionally, since our DualNets have a similar supervised procedure as DER++, this result shows that the DualNets' representation learning and fast adaptation mechanism are beneficial to continual learning.
\blue{Lastly, we observe a similar trend on the large scale ImageNet benchmark where DualNets achieve better evaluation metrics compared to the vanilla strategies of ER and DER++.}

\subsubsection{Ablation Study of the Slow Learner Objectives and Optimizers}\label{sec:exp-slow-obj}
\begin{table*}[t] 
\centering
\caption{DualNet's performance under different slow learner objective and optimizers on the Split miniImageNet-TA benchmark}\label{tab:slow-obj}
\begin{tabular}{@{}lcccccc@{}}
\toprule
\multirow{2}{*}{DualNet} &             & SGD &              &              & Look-ahead &              \\ \cmidrule{2-7}
&\ACC        &\FM           &\LA          &\ACC         &\FM                  &\LA          \\ \midrule
Barlow Twins~\cite{zbontar2021barlow}         & \blue{70.99{\tiny $\pm$1.01}} & \blue{\textbf{3.26{\tiny $\pm$0.53}}}   & \blue{70.93{\tiny $\pm$0.99}} & {\bf 73.20{\tiny $\pm$0.68}} & {\bf 3.86{\tiny $\pm$1.01}}          & {\bf 74.12{\tiny $\pm$0.12}} \\
SimCLR~\cite{chen2020simple}              & {\bf 71.49{\tiny $\pm$1.01}}            & {4.23{\tiny $\pm$0.46}}              &  {\bf 72.64{\tiny $\pm$1.20}}            & 72.13{\tiny $\pm$0.44}        &  4.13{\tiny $\pm$0.52} &    73.09{\tiny $\pm$0.16}           \\
SimSiam~\cite{chen2021exploring} & 70.55{\tiny $\pm$0.98} & 4.93{\tiny $\pm$1.31} & 71.90{\tiny $\pm$0.65} & 71.94{\tiny $\pm$0.64} & 4.21{\tiny $\pm$0.28} & 72.93{\tiny $\pm$0.38} \\
BYOL~\cite{grill2020bootstrap} & 69.76{\tiny $\pm$2.12} & {4.23{\tiny $\pm$1.41}} & 70.33{\tiny $\pm$0.87} & 71.73{\tiny $\pm$0.47} & {\bf 3.96{\tiny $\pm$0.62}} & 72.06{\tiny $\pm$0.28} \\
Classification       &  68.50{\tiny $\pm$1.67} & 5.53{\tiny $\pm$1.67}          & {\bf 72.93{\tiny $\pm$1.10}}      & 70.96{\tiny $\pm$1.08}      & 6.33{\tiny $\pm$0.28}        & 73.92{\tiny $\pm$1.14} \\ \midrule
\multirow{2}{*}{DualNet++} &             & SGD &              &              & Look-ahead &              \\ \cmidrule{2-7}
&\ACC        &\FM           &\LA          &\ACC         &\FM                  &\LA          \\ \midrule
Barlow Twins~\cite{zbontar2021barlow}         & \blue{71.10{\tiny $\pm$1.03}} & \blue{\bf 3.03{\tiny $\pm$0.97}} & \blue{70.96{\tiny $\pm$1.12}} & {\bf 74.24{\tiny $\pm$0.95}} & {\bf 2.83{\tiny $\pm$0.71}} & \bf{74.11{\tiny $\pm$0.30}} \\
SimCLR~\cite{chen2020simple}             & {\bf 71.16{\tiny $\pm$0.72}} & 3.13{\tiny $\pm$1.33}  & 71.96{\tiny $\pm$0.77}  & 72.14{\tiny $\pm$0.81} &  4.12{\tiny $\pm$0.28}   & 73.33{\tiny $\pm$0.58}         \\
SimSiam~\cite{chen2021exploring} & 69.56{\tiny $\pm$0.11}  & 4.53{\tiny $\pm$0.54} & 71.66{\tiny $\pm$1.01} & 71.99{\tiny $\pm$1.33}  & 4.13{\tiny $\pm$0.45} & 73.01{\tiny $\pm$0.57} \\
BYOL~\cite{grill2020bootstrap} & 69.16{\tiny $\pm$1.12} & 4.24{\tiny $\pm$1.21} & 69.93{\tiny $\pm$1.46}  & 71.63{\tiny $\pm$1.12}  & 4.96{\tiny $\pm$1.88} & 72.73{\tiny $\pm$0.77}  \\
Classification       &  69.83{\tiny $\pm$0.36} & 5.26{\tiny $\pm$0.54}          & \bf{72.20{\tiny $\pm$0.66}}      & 70.96{\tiny $\pm$0.52}      & 5.10{\tiny $\pm$0.57}        & 72.96{\tiny $\pm$0.86} \\

\bottomrule
\end{tabular}
\end{table*}

We now study the effects of the slow learner's objective and optimizer on the final performance of DualNets by considering several objectives to train the slow learner. First, we consider the \emph{classification loss} to train the slow net, which reduces DualNet's representation learning to only supervised learning. Second, we consider various contrastive SSL losses, including SimCLR~\cite{chen2020simple}, SimSiam~\cite{chen2021exploring}, and BYOL~\cite{grill2020bootstrap}. In this setting, DualNets' slow representation involves a direct optimization of a SSL loss and an indirect classification loss backpropagated via the fast learner.

We consider the Split miniImageNet-TA and TF benchmark with 50 memory slots per task and optimize each objective using the SGD and Look-ahead optimizers. Tab.~\ref{tab:slow-obj} reports the result of this experiment. 
In general, we observe that SSL objectives achieve a better performance than the classification loss. Moreover, the Look-ahead optimizer consistently improves the performances on all objectives compared to the SGD optimizer. This result shows that the DualNets design is general and can work well with different slow learner's objectives.
Interestingly, when using the Look-ahead optimizer, we observe a correlation between the SSL losses in DualNets with their performances in the standard SSL scenario~\cite{zbontar2021barlow}. This result suggests that DualNets can take advantage of future SOTA SSL losses to further improve the performance.

\subsubsection{Ablation Study of Self-Supervised Learning Iterations}\label{sec:abla-iter}
\begin{figure*}[t]
\captionsetup[subfigure]{justification=Centering}
\begin{subfigure}{\linewidth}
    \centering
    \includegraphics[height=0.4in,width=5.2in]{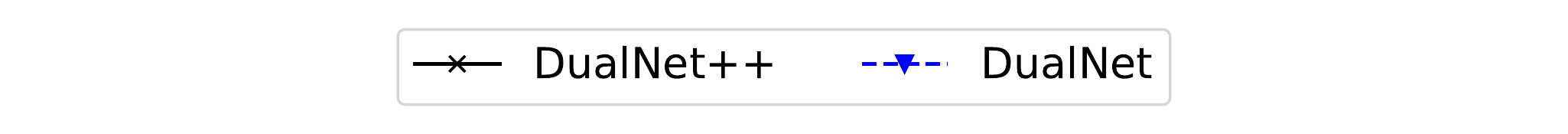}
\end{subfigure}

\centering
\subcaptionbox{}{\includegraphics[width=2.2in]{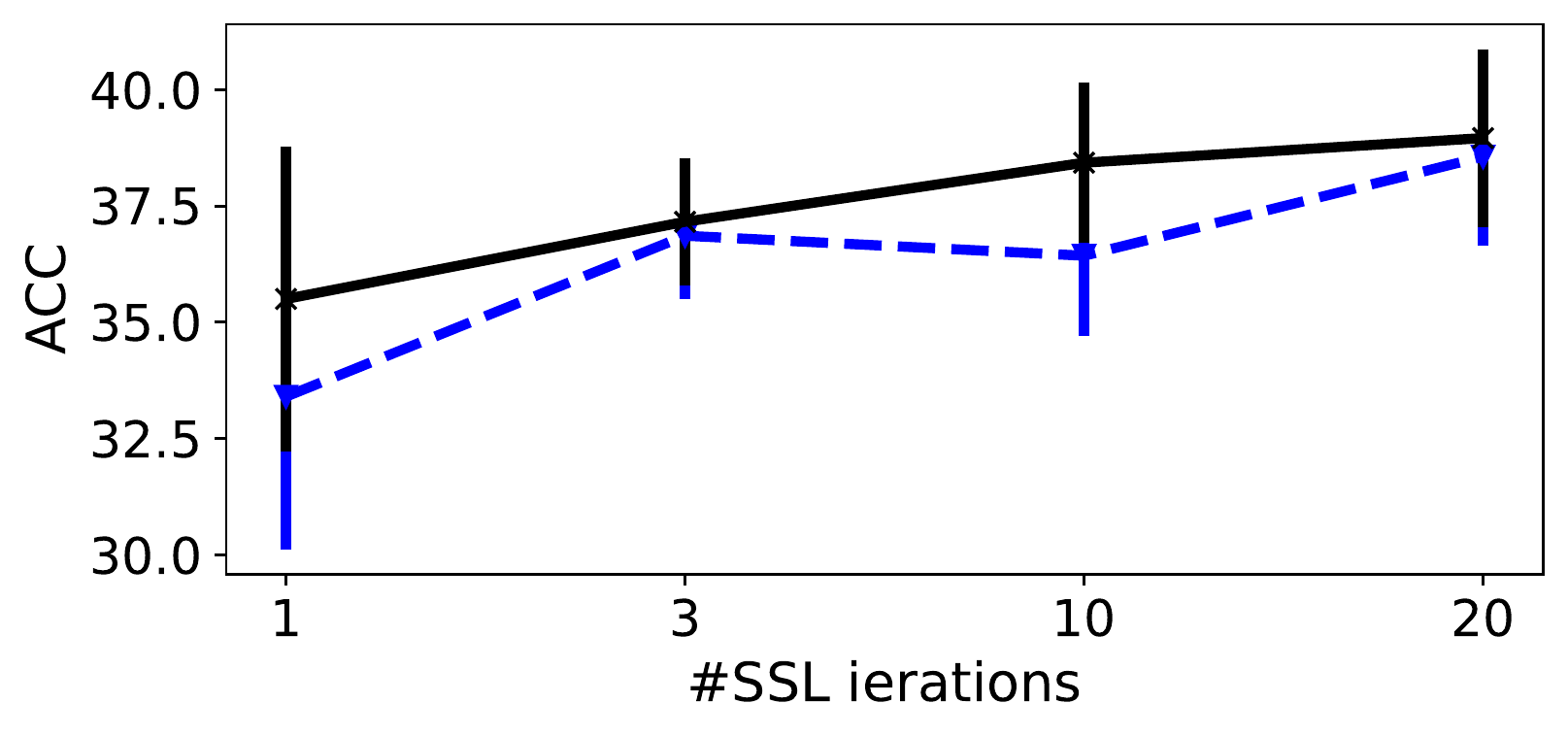}}
\subcaptionbox{Split miniImageNet-TA}{\includegraphics[width=2.2in]{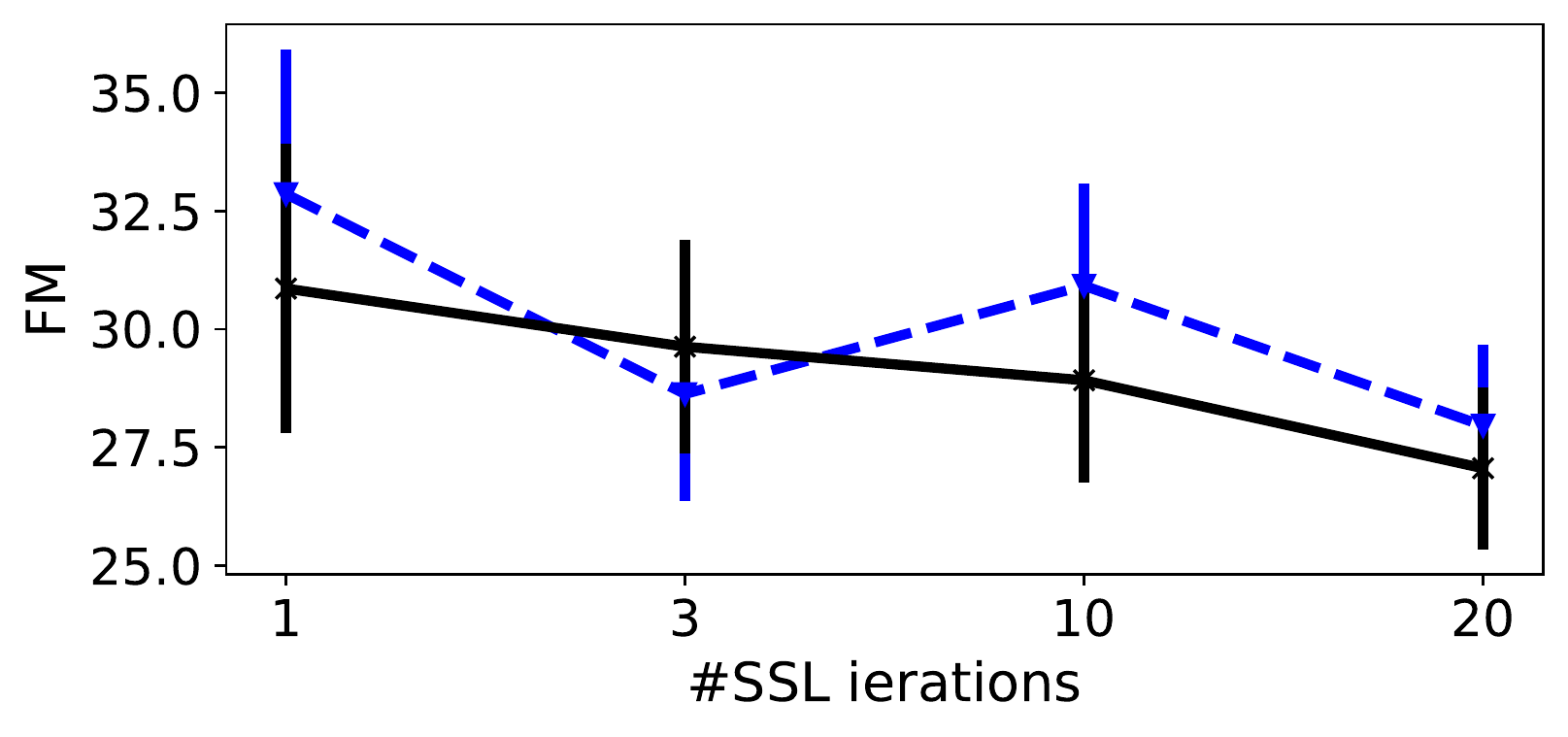}}
\subcaptionbox{}{\includegraphics[width=2.2in]{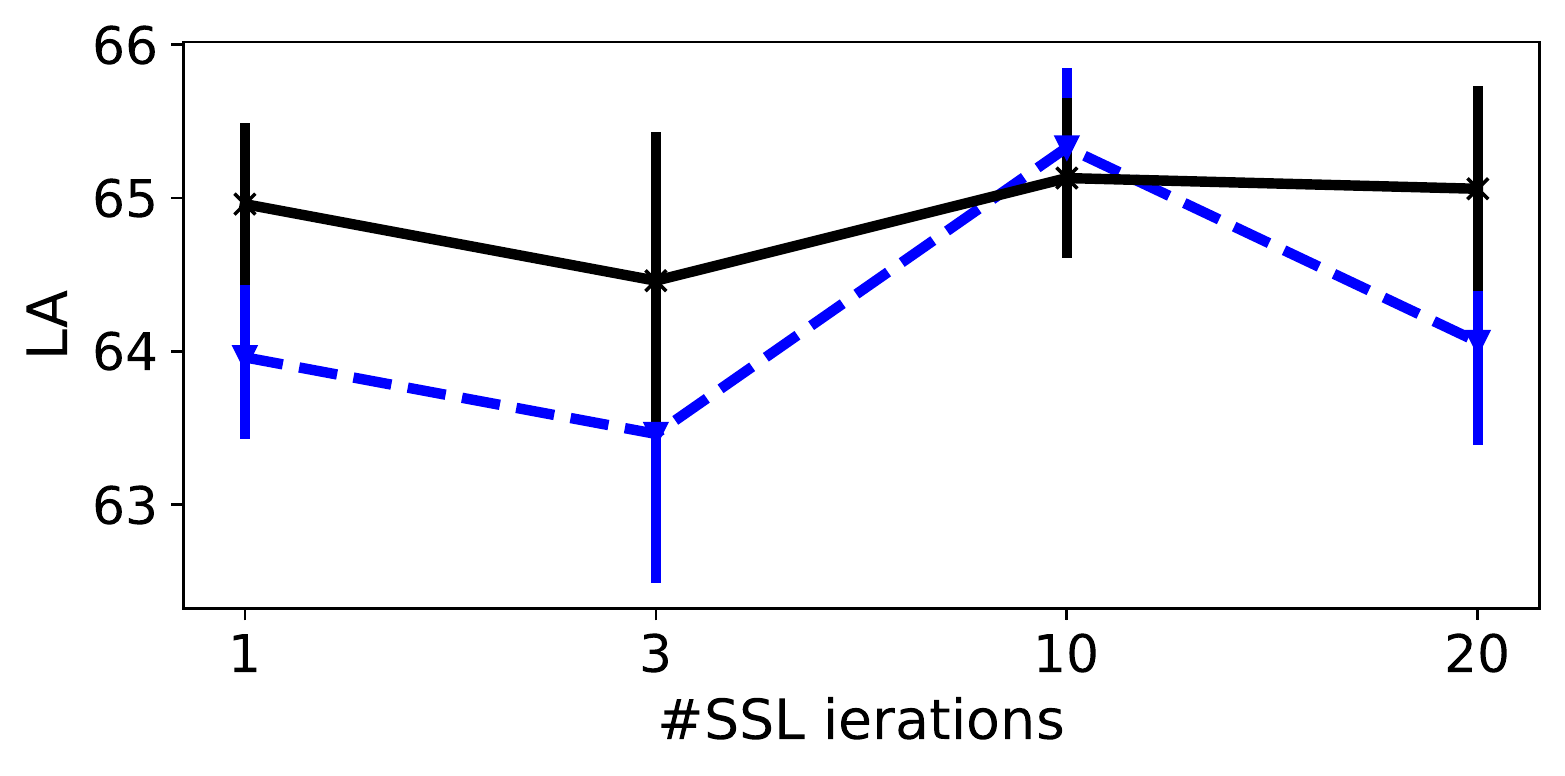}}

\subcaptionbox{}{\includegraphics[width=2.2in]{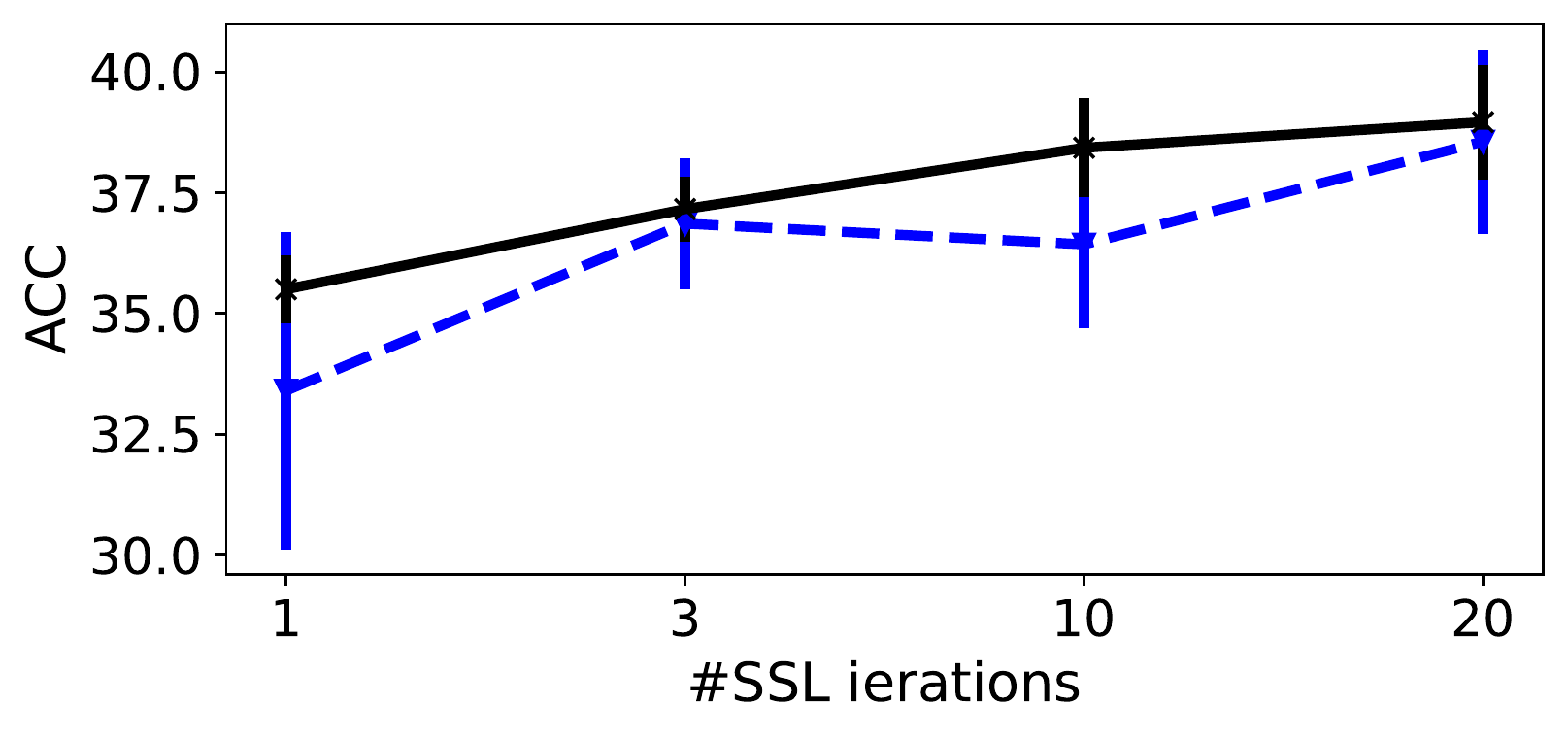}}
\subcaptionbox{Split miniImageNet-TF}{\includegraphics[width=2.2in]{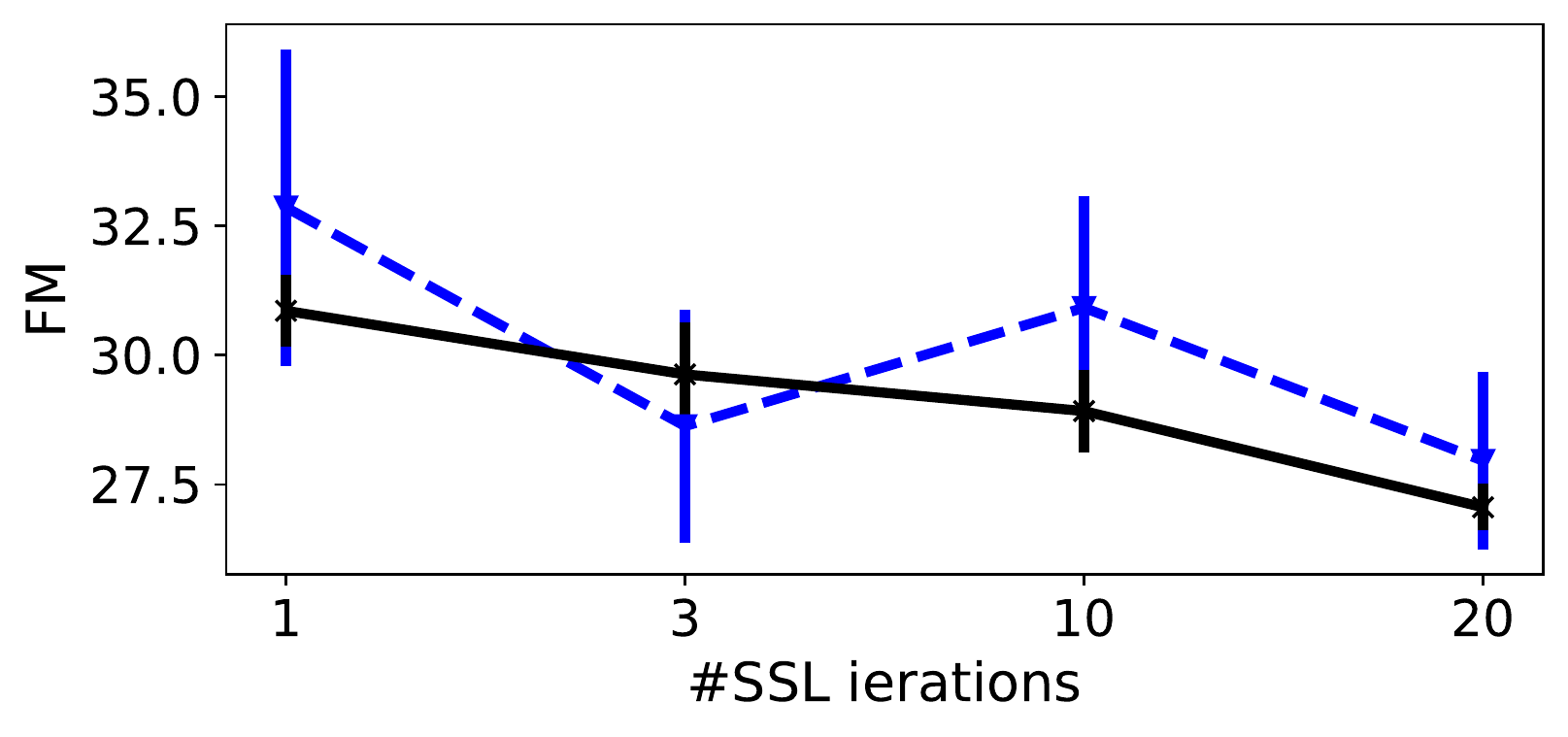}}
\subcaptionbox{}{\includegraphics[width=2.2in]{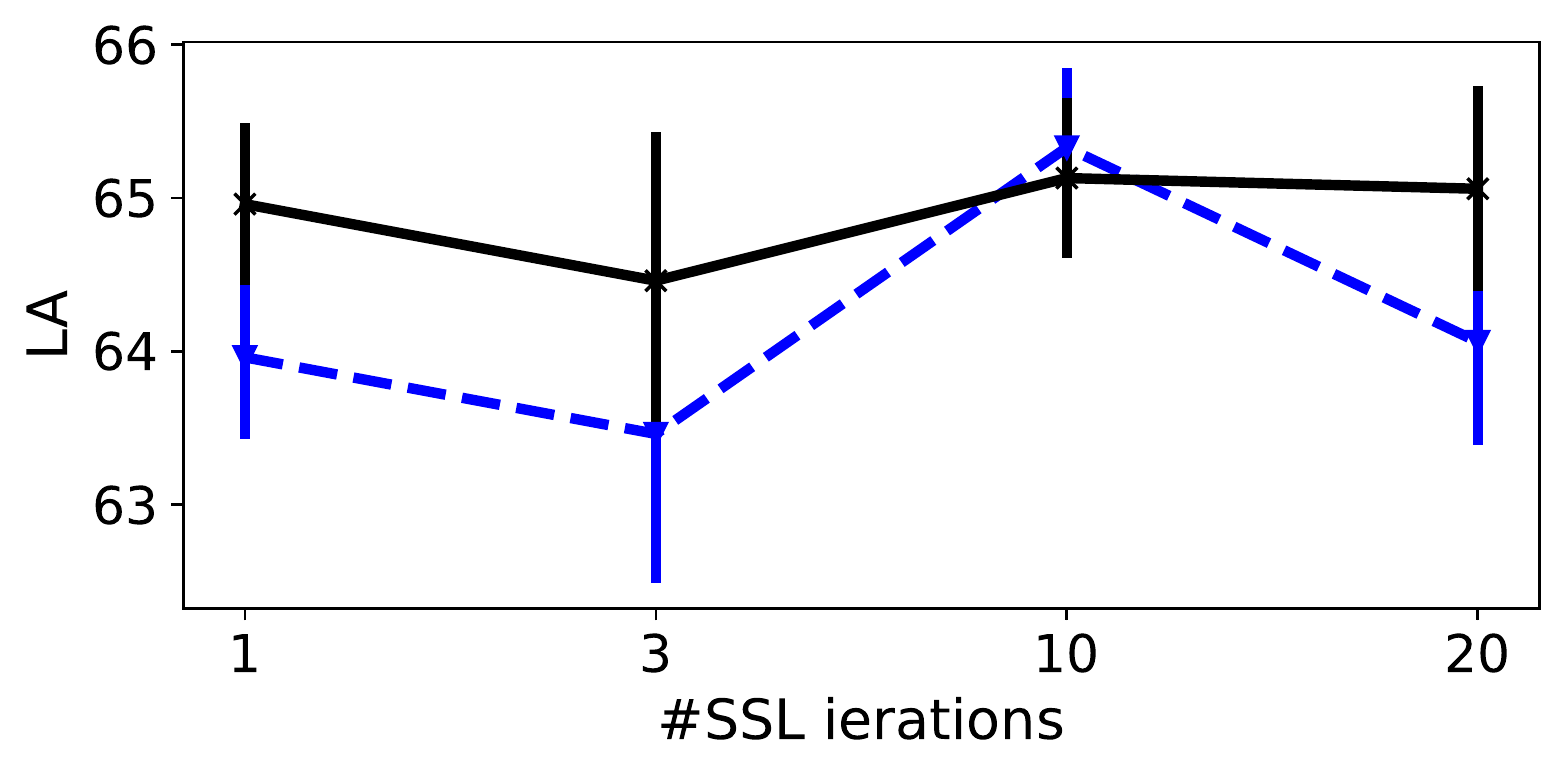}}

\caption{Performance of DualNet and DualNet++ with different self-supervised learning iterations $n$.}
\label{fig:ssl_curve}
\end{figure*}

We now investigate DualNet's performances with different SSL optimization iterations $n$. Small values of $n$ indicate there is little to no delay of labeled data from the continuum, and the fast learner has to query the slow learner's representation continuously. On the other hand, larger $n$ simulates the situations where labeled data are delayed, which allows the slow learner to train its SSL objective for more iterations between each query from the fast learner. In this experiment, we gradually increase the SSL training iterations between each supervised update by varying from $n=1$ to $n=20$.

We run the experiments on both the Split miniImageNet benchmarks under the TA and TF settings. Fig.~\ref{fig:ssl_curve} reports the result of DualNet and DualNet++ in this scenario.
In general, we observe that in all cases, the average accuracy \ACC increases as more SSL iterations are allowed. The same conclusion also holds for the \FM and \LA metrics, although there are small fluctuations at $n=3$ and $n=10$. Moreover, DualNet++ is more stable than DualNet in this experiment by having smaller variance across different runs. We can conclude that both DualNet and DualNet++ are highly scalable with the number of SSL training iterations. This promising result demonstrates the DualNets' potential to be deployed in real-world continual learning scenarios where labeled data is delayed~\cite{diethe2019continual}, which allows the slow learner to learn in the background.

\subsubsection{Ablation Study of DualNets Components} \label{sec:abla-dual}

\begin{table}[t]
\centering
\caption{Evaluation of DualNet's slow learner on the Split miniImageNet TA and TF benchmarks}
\label{tab:fast}
\begin{tabular}{lccc}
\toprule
\multirow{2}{*}{DualNet} & \multicolumn{3}{c}{Split miniImageNet-TA} \\ \cmidrule{2-4}
 & \ACC & \FM & \LA \\ \midrule
Slow + Fast Nets & {\bf 73.20{\tiny $\pm$0.68}} & {\bf 3.86{\tiny $\pm$1.01}} & {\bf 74.12{\tiny $\pm$0.12}} \\
Slow Net & 68.33{\tiny $\pm$0.57} & 5.12{\tiny $\pm$0.78} & 69.20{\tiny $\pm$0.32} \\ \midrule
\multirow{2}{*}{DualNet} & \multicolumn{3}{c}{Split miniImageNet-TF} \\ \cmidrule{2-4}
 & \ACC & \FM & \LA \\ \midrule
Slow + Fast Nets & {\bf 36.86{\tiny $\pm$1.36}} & {\bf 28.63{\tiny $\pm$2.26}} & {\bf 63.46{\tiny $\pm$1.97}} \\
Slow Net & 27.30{\tiny $\pm$0.25} & 34.60{\tiny $\pm$1.12} & 59.70{\tiny $\pm$1.26} \\ \bottomrule
\end{tabular}
\vspace*{-0.1in}
\end{table}

Compared to the standard ER strategy with soft labels~\cite{van2018generative,buzzega2020dark}, DuelNets introduce an additional fast learner and a representation learning phase. In this experiment, we investigate the contribution of the fast learner on the Split miniImageNet benchmark, both the TA and TF settings. We create a variant, \emph{Slow Learner}, that uses only a ResNet backbone to optimize both the supervised and SSL losses. Tab.~\ref{tab:fast} reports the result of this experiment.
We can see that the slow learner variant binds both representation types into the same backbone and performs significantly worse than the original DualNet in both scenarios. \blue{It is worth noting that using only the slow learner to learn both representations is the strategy in concurrent works~\cite{cha2021co2l,bhat2022task}.} This promising result corroborates with our motivation in Sec.~\ref{sec:intro} that it is more beneficial to separate the two representations into two distinct systems.

\subsubsection{Semi-Supervised Continual Learning Setting} \label{sec:semi}
\begin{table}[t]
\centering
\caption{Evaluation metrics on the Split miniImageNet-TA benchmarks under the semi-supervised setting, where $\rho$ denotes the fraction of data that is labeled}
\label{tab:semi}
{\begin{tabular}{@{}lccc@{}}
\toprule
\multirow{2}{*}{Method} & \multicolumn{3}{c}{$\rho = 4\%$} \\ \cmidrule(l){2-4} 
                        &\ACC         &\FM         &\LA  \\ \midrule
ER                      & 40.26{\tiny $\pm$0.27} & 2.76{\tiny $\pm$1.24} & 34.69{\tiny $\pm$1.09} \\
DER++                     & 39.46{\tiny $\pm$0.42} & 3.92{\tiny $\pm$0.46} & 33.02{\tiny $\pm$1.01} \\
CTN                     & 47.86{\tiny $\pm$0.66} & 2.96{\tiny $\pm$1.01} & 44.56{\tiny $\pm$0.29}  \\ 
Cassle                     & 43.70{\tiny $\pm$2.11} & 4.02{\tiny $\pm$1.99} & 44.70{\tiny $\pm$1.09}  \\ 
\midrule
DualNet                 & \underline{51.79{\tiny $\pm$0.58}}  & \underline{2.66{\tiny $\pm$0.97}} & \textbf{45.77{\tiny $\pm$0.37}}   \\ 
DualNet++                 & \textbf{53.13{\tiny $\pm$0.19}}  & {\bf 1.86{\tiny $\pm$0.49}} & \underline{44.79{\tiny $\pm$0.21}} \\ \midrule
\multirow{2}{*}{Method} & \multicolumn{3}{c}{$\rho = 10\%$} \\ \cmidrule(l){2-4} 
                        &\ACC         &\FM         &\LA  \\ \midrule
ER                      & 41.66{\tiny $\pm$2.72} & 6.80{\tiny $\pm$2.07} & 42.33{\tiny $\pm$1.51} \\
DER++                     & 44.56{\tiny $\pm$1.41} & 4.55{\tiny $\pm$0.66} & 43.03{\tiny $\pm$0.71} \\
CTN                     & 49.80{\tiny $\pm$2.66} & 3.96{\tiny $\pm$1.16} & 47.76{\tiny $\pm$0.99}  \\ 
Cassle                     &  45.80{\tiny $\pm$1.86} & 5.90{\tiny $\pm$0.96} & 46.66{\tiny $\pm$1.07}  \\ 
\midrule
DualNet                 & \underline{54.03{\tiny $\pm$2.88}}  & \underline{3.46{\tiny $\pm$1.17}} & \underline{49.96{\tiny $\pm$0.17}}   \\ 
DualNet++                 & \textbf{58.03{\tiny $\pm$0.99}}  & {\bf 2.16{\tiny $\pm$0.59}} & {\bf 53.56{\tiny $\pm$0.11}} \\ \midrule
\multirow{2}{*}{Method} & \multicolumn{3}{c}{$\rho = 25\%$} \\ \cmidrule(l){2-4} 
                        &\ACC         &\FM         &\LA  \\ \midrule
ER                       & 50.13{\tiny $\pm$2.19} & 6.76{\tiny $\pm$1.51} & 51.90{\tiny $\pm$2.16} \\
DER++                    & 51.63{\tiny $\pm$1.11} & 6.03{\tiny $\pm$1.46} & 52.36{\tiny $\pm$0.55} \\
CTN                   & 55.90{\tiny $\pm$0.86} & 3.84{\tiny $\pm$0.32} & 55.69{\tiny $\pm$0.98} \\ 
Cassle                     & 48.33{\tiny $\pm$2.27} & 8.26{\tiny $\pm$0.72} & 52.10{\tiny $\pm$2.04}  \\ \midrule
DualNet                  &   \underline{62.80{\tiny $\pm$2.40}}           &  \underline{3.13{\tiny $\pm$0.99}} & \underline{59.60{\tiny $\pm$1.87}}              \\ 
DualNet++                 &   {\bf 63.96{\tiny $\pm$1.22}}           &  {\bf 2.20{\tiny $\pm$0.32}} & {\bf 60.66{\tiny $\pm$1.02}}              \\
\bottomrule
\end{tabular}}
\vspace*{-0.1in}
\end{table}

In real-world continual learning scenarios, there exist abundant unlabeled data, which are costly and even unnecessary to label entirely. Therefore, a practical continual learning system should be able to improve its representation using unlabeled samples while waiting for the labeled data. To test the performance of existing methods in such scenarios, we create a \emph{semi-supervised continual learning} benchmark, where the data stream contains both labeled and unlabeled data. For this, we consider the Split miniImageNet-TA benchmark but provide labels randomly to a fraction ($\rho$) of the total samples, which we set to be \blue{$\rho=4\%$, $\rho=10\%$ and $\rho=25\%$}. The remaining samples are unlabeled and cannot be processed by the baselines we have considered so far. In contrast, such samples can go directly to the DualNet's slow learner to improve its representation while the fast learner stays inactive. Other configurations remain the same as the experiment in Sec.~\ref{sec:main-results}. \blue{For this experiment, we also consider and adapt Cassle~\cite{fini2022self}, a method that introduces SSL to continual learning. Note that originally, Cassle was developed for a two phase learning: a pre-training the Cassle loss using the unlabeled data; and a linear evaluation phase. Therefore, we adapt Cassle to our setting by training Cassle simultaneously with supervised learning on the data stream.}

Tab.~\ref{tab:semi} shows the results of this experiment. Under the limited labeled data regimes (e.g. $\rho=4\%$ or $\rho=10\%$), the results of ER and DER++ drop significantly. Meanwhile, CTN can still maintain competitive performances thanks to additional information from the task identifiers, which remains untouched. On the other hand, both DualNet and DualNet++ can efficiently leverage the unlabeled data to improve its performance and outperform other baselines, even CTN. 
\blue{We also observe that Cassle achieves competitive performances in the low data regimes thanks to its strong SSL loss. However, since Cassle does not store past samples, its performance quickly degraded and is outperformed by other baselines when more labeled samples are available ($\rho=25\%$).}
\blue{Lastly, DualNet++ consistently outperforms DualNet, and the gaps are larger as the labeled samples are more limited.} This gap is attributed to the dropout layers in DualNet++, which improve the fast learner's ability to use the slow features and to better perform supervised learning.
Overall, the result demonstrates DualNets potential to work in a real-world environment, where data is partially labeled.

\begin{table}[ht]
\centering
\caption{Performance of the Offline model under different configuration on the Split miniImageNet-TA benchmark, * denotes the method is trained in the continual learning setting - for reference} \label{tab:offline}
\begin{tabular}{lcccc}
\toprule
Architecture                    & Method  & Loss   & Data Aug & ACC          \\ \midrule
\multirow{4}{*}{Fast+Slow nets} & DualNet* & SL+SSL & Yes      & 73.20{\tiny $\pm$0.68} \\
                                & Offline & SL     & No       & 75.83{\tiny $\pm$1.07} \\
                                & Offline & SL     & Yes      & {77.63{\tiny $\pm$0.48}} \\
                                & Offline & SL+SSL & Yes      & {\bf 77.98{\tiny $\pm$0.16}} \\ \midrule
\multirow{2}{*}{Slow net}       & Offline & SL     & No       & 71.15{\tiny $\pm$2.95} \\
                                & Offline & SL     & Yes      & 75.46{\tiny $\pm$0.97} \\ \bottomrule
\end{tabular}
\end{table}

\subsubsection{DualNet's Upper Bound}
In our work, there are three factors affecting the DualNets' upper bound: (i) model architecture: slow net (standard backbone) versus fast and slow nets (DualNets); (ii) training loss: supervised learning loss (SL) or supervised and self-supervised learning losses (SL+SSL); and (iii) data augmentation. As a result, we believe that an upper bound of DualNet is a model having all three factors as DualNet (has fast and slow learners, optimized both SL and SSL losses with data augmentation) and is trained offline. The offline model has access to all tasks' data to simultaneously optimizes both the SL and SSL losses, which are backpropagated through both learners. Here we consider the offline model trained up to five epochs.

We explore different combinations of the aforementioned factors to train an Offline model on the Split miniImageNet-TA benchmark and report the result in Tab.~\ref{tab:offline}. Note that the configuration of Slow Net + Offline + SL + no data augmentation is the previous result reported in \cite{pham2021contextual}. Our argued upper bound for DualNet has the following configuration: Fast + Slow nets + Offline + SL + SSL + data augmentation. The result confirms the upper bound of DualNet. Moreover, in the offline training with all data, the SSL only contributes a minor improvement to the SL. However, in continual learning, SSL is more beneficial because its representation does not depend on the class label, and therefore more resistant to catastrophic forgetting when old task data is limited.

\begin{table*}[t]
\centering
\caption{Details of the CTrL benchmark streams built from five common datasets: CIFAR-10~\cite{krizhevsky2009learning}, MNIST~\cite{lecun1998gradient}, DTD~\cite{cimpoi2014describing}, F-MNIST (Fashion MNIST)~\cite{xiao2017fashion}, and SVHN~\cite{netzer2011reading}}
\label{tab:ctrl}
\begin{tabular}{llcccccc}
\toprule
\textbf{Stream} & Configuration & T1 & T2 & T3 & T4 & T5 & T6 \\ \midrule
\multirow{3}{*}{\textbf{$\mc S^+$}} & \textbf{Dataset} & CIFAR-10 & MNIST & DTD & F-MNIST & SVHN & CIFAR-10 \\
 & \textbf{\# Train samples} & 4000 & 400 & 400 & 400 & 400 & 400 \\
 & \textbf{\# Val. samples} & 2000 & 200 & 200 & 200 & 200 & 200 \\ \midrule
\multirow{3}{*}{\textbf{$\mc S^-$}} & \textbf{Dataset} & CIFAR-10 & MNIST & DTD & F-MNIST & SVHN & CIFAR-10 \\
 & \textbf{\# Train samples} & 400 & 400 & 400 & 400 & 400 & 4000 \\
 & \textbf{\# Val. samples} & 200 & 200 & 200 & 200 & 200 & 2000 \\ \midrule
\multirow{3}{*}{\textbf{$\mc S^{\mathrm{in}}$}} & \textbf{Dataset} & R-MNIST & CIFAR-10 & DTD & F-MNIST & SVHN & R-MNIST \\
 & \textbf{\# Train samples} & 4000 & 400 & 400 & 400 & 400 & 50 \\
 & \textbf{\# Val. samples} & 2000 & 200 & 200 & 200 & 200 & 30 \\ \midrule
\multirow{3}{*}{\textbf{$\mc S^{\mathrm{out}}$}} & \textbf{Dataset} & CIFAR-10 & MNIST & DTD & F-MNIST & SVHN & CIFAR-10 \\
 & \textbf{\# Train samples} & 4000 & 400 & 400 & 400 & 400 & 400 \\
 & \textbf{\# Val. samples} & 2000 & 200 & 200 & 200 & 200 & 200 \\ \midrule
\multirow{3}{*}{\textbf{$\mc S^{\mathrm{pl}}$}} & \textbf{Dataset} & MNIST & DTDd & F-MNIST & SVHN & CIFAR-10 & - \\
 & \textbf{\# Train samples} & 400 & 400 & 400 & 400 & 4000 & - \\
 & \textbf{\# Val. samples} & 200 & 200 & 200 & 200 & 2000 & - \\ \bottomrule
\end{tabular}
\end{table*}

\subsection{Batch Continual Learning Experiments}\label{sec:batch-cl}
We now consider the \emph{Batch Continual Learning setting}~\cite{kirkpatrick2017overcoming}, where all data samples of a task arrive at each continual learning step. As a result, the model is allowed to train on these samples for multiple epochs before moving on to the next task. Thus, most methods do not suffer from the difficulties of training deep neural networks online and focus only on preventing catastrophic forgetting~\cite{buzzega2020dark}.

\subsubsection{Setups}
\textbf{The CTrL Benchmark } In the batch learning setting, we focus on exploring DualNet's ability to facilitate knowledge transfer in complex continual learning scenarios. To this end, we consider the CTrL benchmark~\cite{veniat2020efficient}, which was carefully designed to access the model's ability to selectively transfer knowledge while avoiding catastrophic forgetting. Before introducing the CTrL streams, we briefly summarize the concept of related tasks used in CTrL.

We denote a continual learning task as $\mc T$. Then, CTrL introduces four variants of $\mc T$ as: (1) $\mc T^-$: a task whose data is sampled from the same distribution as $\mc T$ but has a much smaller number of training samples; (2) $\mc T^+$ is similar to $\mc T^-$ but has much more training samples than $\mc T$; (3) $\mc T'$ is similar to $\mc T$ but has a different input distribution, i.e., different background colors; and (4) $\mc T''$ is similar to $\mc T$ but has a different output distribution, i.e., the label order is randomly permuted. In addition, there are no relationships between two tasks that have different subscripts.
With this notation, the CTrL benchmarks introduce five continual learning streams to evaluate five basic knowledge transfer abilities comprehensively.

In the \textbf{$\mc S^- = \{\mc T_1^+, \mc T_2, \mc T_3, \mc T_4, \mc T_5, \mc T_1^- \}$} stream, the last task is similar to the first one but it has much smaller training samples. Therefore, successful methods must remember and transfer the knowledge after learning four unrelated tasks.

In the \textbf{$\mc S^+ = \{\mc T_1^-, \mc T_2, \mc T_3, \mc T_4, \mc T_5, \mc T_1^+ \}$} stream, the first task is similar but has much smaller data than the last one, which requires the model to remember and update the knowledge after learning irrelevant tasks.

The \textbf{$\mc S^{\mathrm{in}} = \{\mc T_1, \mc T_2, \mc T_3, \mc T_4, \mc T_5, \mc T_1' \}$} and \mbox{\textbf{$\mc S^{\mathrm{out}} = \{\mc T_1^, \mc T_2, \mc T_3, \mc T_4, \mc T_5, \mc T_1'' \}$}} streams require the model to learn representations that are useful to either input or output distribution shifts.

Lastly, the \textbf{$\mc S^{\mathrm{pl}} = \{\mc T_1, \mc T_2, \mc T_3, \mc T_4, \mc T_5 \}$} stream test the model's ability to learn unrelated tasks with the potential interference from unrelated features.
Tab.~\ref{tab:ctrl} provides the details of each stream in the CTrL benchmark.  Interestingly, the $\mc S^-, \mc S^+, \mc S^{\mathrm{in}}$ and $\mc S^{\mathrm{out}}$ streams contain one unrelated task from the DTD dataset~\cite{cimpoi2014describing}, which has very different visual features from the remaining tasks (see Tab.~\ref{tab:ctrl}).
Therefore, we believe the CTrL benchmark can access DualNets' ability to selective transfer useful knowledge under the presence of negative transfer.

\textbf{Baselines } We follow the experimental setting in~\cite{ostapenko2021continual} and compare DualNets with a suite of competitive baselines. First, we consider static architecture approaches of \textbf{EWC}~\cite{kirkpatrick2017overcoming} and \textbf{O-EWC}~\cite{huszar2017quadratic} (online EWC), which use a quadratic regularizer to penalize changes to important parameters of previous tasks according to the Fisher information. The original EWC~\cite{kirkpatrick2017overcoming} maintains an estimate of the Fisher information matrix for each task, while O-EWC maintains a moving average of the parameters importance. Next, we include experience replay \textbf{ER}~\cite{chaudhry2019tiny}, dark experience replay \textbf{DER++}~\cite{buzzega2020dark}, \blue{\textbf{CLSER}~\cite{arani2021learning}}, and the naive \textbf{Finetune} strategy that trains a single model without any continual learning strategies. We also consider the task-free and task-aware variants of these baselines.

Second, we consider a suite of dynamic architecture approaches. The \textbf{Independent}~\cite{lopez2017gradient} baseline trains a separate model for each task. \text{HAT}~\cite{hat-cl} proposes to learn hard attention masks to gate the backbone network, which prevents catastrophic forgetting. \textbf{SG-F}~\cite{mendez2020lifelong} proposes a task-specific structural network that learns to update existing modules, combine modules, and add new modules. \textbf{MNTDP}~\cite{veniat2020efficient} organizes the backbone network into modules and efficiently searches the path configuration to connect the modules to solve a given task. Lastly, \textbf{LMC}~\cite{ostapenko2021continual}, a recent dynamic architecture method that proposes to equip each module with a local structural component to predict its relevance for a given input, which does not require task identifier at test time~\footnote{LMC still requires task identifiers during training, which we consider as a task-aware method.} and provide a more systematic strategy to expand and search the layout over modules. 

\textbf{Training } We follow the training procedure provided in~\cite{ostapenko2021continual} for a fair comparison. Particularly, for each task, we train each method over 100 epochs using the Adam optimizer~\cite{kingma2015adam}. We use a weak data augmentation of random flipping and random cropping for both the supervised learning phase of all methods and the self-supervised learning phase of DualNet. Furthermore, since DualNets are allowed to train for 100 epochs, it is sufficient to use the standard SGD to optimize the SSL loss. Our preliminary experiments show that in this setting, the Look-ahead and SGD optimizers achieve similar results. \blue{We also observe that CLSER has several hyper-parameters, and there are not any guidelines to set them for new applications. Thus, we considered the hyper-parameter configurations provided in~\cite{arani2021learning} and reported the one with the best \ACC.}
\blue{For the backbone network, we consider two variants: a full ResNet18 trained from scratch or pre-trained on ImageNet.} 

Regarding the model complexity, we anchor on the final model of LMC, the state-of-the-art method on this benchmark, to calculate the total parameters used. Then, we select the replay buffer size of each method so that their total parameters~\footnote{total parameters = model parameters + memory parameters in floating point numbers.} equals to LMC. We repeat each experiment five times and report the average \ACC and \BWT at the end of learning. The metrics are defined in Sec.~\ref{sec:benchmark}.

\begin{table*}[t]
\centering
\caption{Evaluation metrics on the CTrL benchmark, we report the average accuracy \ACC and backward transfer \BWT at the end of training. The (H) suffix denotes the hard module selection of LMC, $^*$ denotes methods that use more parameters. We highlight the methods with best mean metrics in bold, and underline the second best methods}
\label{tab:ctrl-result}
{\begin{tabular}{lcccccccccc}
\toprule
\multirow{2}{*}{Method}         & \multicolumn{2}{c}{$\mc S^-$}            & \multicolumn{2}{c}{$\mc S^+$}    & \multicolumn{2}{c}{$\mc S^{\mathrm{in}}$}    & \multicolumn{2}{c}{$\mc S^{\mathrm{out}}$}  & \multicolumn{2}{c}{$\mc S^{\mathrm{pl}}$}  \\ \cmidrule{2-11}
      & \ACC                   & \BWT         & \ACC          & \BWT          & \ACC           & \BWT          & \ACC           & \BWT         & \ACC          & \BWT         \\ \midrule
\multicolumn{11}{c}{Task-free \& Train from scratch}                                                                                                                                 \\ \midrule
Finetune   & 47.5{\tiny $\pm$ 1.5}          & -14.9{\tiny $\pm$1.4} & 31.4{\tiny $\pm$3.7} & -29.3{\tiny $\pm$3.8}  & 39.7{\tiny $\pm$5.0}  & -23.9{\tiny $\pm$5.7}  & 45.4{\tiny $\pm$4.0}  & -15.5{\tiny $\pm$3.7} & 29.1{\tiny $\pm$3.1} & -29.2{\tiny $\pm$3.2} \\
Finetune-L  & 52.1{\tiny $\pm$1.4}          & -15.7{\tiny $\pm$1.7} & 38.2{\tiny $\pm$3.2} & -25.08{\tiny $\pm$3.3} & 49.3{\tiny $\pm$2.0}  & -18.4{\tiny $\pm$2.0}  & 49.3{\tiny $\pm$2.1}  & -18.4{\tiny $\pm$2.0} & 37.1{\tiny $\pm$2.1} & -26.0{\tiny $\pm$2.2} \\
ER         & 61.5{\tiny $\pm$0.5}          &  {-3.1{\tiny $\pm$1.2}}  & {59.9{\tiny $\pm$0.5}} & {0.2{\tiny $\pm$0.9}}    & 50.0{\tiny $\pm$1.4}  & -12.6{\tiny $\pm$0.2}  & 51.0{\tiny $\pm$0.5}  & -6.3{\tiny $\pm$1.2}  & {54.6{\tiny $\pm$2.2}} & -5.7{\tiny $\pm$2.3}  \\ 
DER++         & 63.0{\tiny $\pm$0.7}          &  {-2.6{\tiny $\pm$0.4}}  & {59.9{\tiny $\pm$0.9}} & {0.1{\tiny $\pm$0.9}}    & 55.8{\tiny $\pm$2.6}  & -10.5{\tiny $\pm$1.9}  & 55.4{\tiny $\pm$0.6}  & -0.9{\tiny $\pm$0.4}  & {58.1{\tiny $\pm$1.7}} & -3.3{\tiny $\pm$0.6}  \\
CLSER         & 62.1{\tiny $\pm$0.5}          &  {-3.2{\tiny $\pm$0.4}}  & {59.1{\tiny $\pm$1.2}} & {0.1{\tiny $\pm$0.9}}    & 58.0{\tiny $\pm$1.3}  & -7.7{\tiny $\pm$1.5}  & 51.0{\tiny $\pm$0.8}  & -4.9{\tiny $\pm$0.6}  & {57.6{\tiny $\pm$0.4}} & -1.2{\tiny $\pm$1.1} \\
MNTDP   & 41.9{\tiny $\pm$2.5}          & -2.8{\tiny $\pm$0.6}  & 43.2{\tiny $\pm$1.3} & -10.8{\tiny $\pm$2.0}  & 32.7{\tiny $\pm$13.6} & -15.2{\tiny $\pm$13.2} & 37.9{\tiny $\pm$2.7}  & -5.8{\tiny $\pm$3.5}  & 35.1{\tiny $\pm$3.6} & -16.4{\tiny $\pm$4.6} \\
SG-F   & 29.5{\tiny $\pm$3.5}          & -35.3{\tiny $\pm$4.0} & 20.4{\tiny $\pm$4.4} & -39.3{\tiny $\pm$6.7}  & 24.4{\tiny $\pm$5.6}  & -38.7{\tiny $\pm$4.0}  & 30.5{\tiny $\pm$4.5} & -34.0{\tiny $\pm$5.5} & 19.4{\tiny $\pm$1.0} & -41.8{\tiny $\pm$1.6} \\ 
Cassle         & 65.2{\tiny $\pm$0.5}          &  {-4.2{\tiny $\pm$0.5}}  & {60.1{\tiny $\pm$1.5}} & {-0.1{\tiny $\pm$1.7}}    & 56.4{\tiny $\pm$4.0}  & -9.3{\tiny $\pm$4.0}  & 53.4{\tiny $\pm$0.9}  & -9.1{\tiny $\pm$2.1}  & {58.5{\tiny $\pm$0.9}} & -2.9{\tiny $\pm$0.9}  
\\ \midrule
DualNet    & \underline{65.5{\tiny $\pm$0.6}} & \underline{-1.8{\tiny $\pm$0.4}}  & \underline{61.1{\tiny $\pm$0.7}} & \underline{1.3{\tiny $\pm$0.6}}    & \underline{59.1{\tiny $\pm$0.5}}  & \underline{-6.8{\tiny $\pm$0.5}}   & \underline{55.7{\tiny $\pm$0.4}}  & \underline{-0.7{\tiny $\pm$0.6}}  & \underline{58.9{\tiny $\pm$2.2}} & \underline{-1.8{\tiny $\pm$0.1}} \\
DualNet++    & \textbf{68.7{\tiny $\pm$0.7}} & \textbf{-0.1{\tiny $\pm$0.1}}  & \textbf{62.9{\tiny $\pm$0.8}} & \textbf{2.9{\tiny $\pm$1.4}}    & \textbf{61.6{\tiny $\pm$1.1}}  & \textbf{-5.1{\tiny $\pm$0.6}}   & \textbf{57.3{\tiny $\pm$0.9}}  & \textbf{-0.6{\tiny $\pm$0.3}}  & \textbf{59.7{\tiny $\pm$1.2}} & \textbf{0.2{\tiny $\pm$0.4}} \\ \midrule
\multicolumn{11}{c}{Task-free \& Initialized from a pre-trained ImageNet model}  \\ \midrule
ER         & 62.5{\tiny $\pm$0.8}          &  {-2.6{\tiny $\pm$0.8}}  & {62.2{\tiny $\pm$1.3}} & \underline{0.7{\tiny $\pm$1.4}}    & 52.4{\tiny $\pm$3.8}  & -10.9{\tiny $\pm$3.4}  & 51.3{\tiny $\pm$2.2}  & -6.8{\tiny $\pm$1.8}  & {55.2{\tiny $\pm$1.8}} & -4.8{\tiny $\pm$1.2}  \\ 
DER++         & 64.0{\tiny $\pm$0.7}          &  \underline{-2.4{\tiny $\pm$1.3}}  & {63.3{\tiny $\pm$0.8}} & \textbf{0.9{\tiny $\pm$0.9}}    & 57.1{\tiny $\pm$0.9}  & -10.2{\tiny $\pm$0.9}  & 54.6{\tiny $\pm$1.4}  & -4.7{\tiny $\pm$1.1}  & {58.9{\tiny $\pm$2.8}} & -4.5{\tiny $\pm$2.3}  \\ \midrule
DualNet        & \underline{66.7{\tiny $\pm$0.7}}          &  \underline{-2.4{\tiny $\pm$0.2}}  & \underline{67.9{\tiny $\pm$0.2}} & {0.6{\tiny $\pm$0.5}}    & \underline{62.8{\tiny $\pm$0.9}}  & \underline{-6.4{\tiny $\pm$1.1}}  & \underline{55.2{\tiny $\pm$1.0}}  & \underline{-3.2{\tiny $\pm$0.6}}  & \underline{63.9{\tiny $\pm$2.1}} & \underline{-2.7{\tiny $\pm$0.7}}  \\
DualNet++         & \textbf{68.3{\tiny $\pm$0.5}}          &  \textbf{-2.2{\tiny $\pm$0.7}}  & \textbf{68.3{\tiny $\pm$0.4}} & {0.6{\tiny $\pm$0.6}}    & \textbf{63.1{\tiny $\pm$0.7}}  & \textbf{-6.2{\tiny $\pm$1.5}}  & \textbf{55.3{\tiny $\pm$0.6}}  & \textbf{-3.1{\tiny $\pm$0.7}}  & \textbf{64.2{\tiny $\pm$2.4}} & -2.4{\tiny $\pm$0.5}  \\\midrule

\multicolumn{11}{c}{Task-aware \& Train from scratch}  \\ \midrule
Independent$^*$    & 62.7{\tiny $\pm$0.9}         & 0.0           & 63.2{\tiny $\pm$0.8} & 0.0         & 63.1{\tiny $\pm$0.7}  & 0.0         & 63.1{\tiny $\pm$0.7}  & 0.0        & 63.9{\tiny $\pm$0.5} & 0.0        \\ 
EWC        & {62.7{\tiny $\pm$0.7}}          & -3.6{\tiny $\pm$0.9}  & 53.4{\tiny $\pm$1.8} & -2.3{\tiny $\pm$0.4}   & {56.3{\tiny $\pm$2.5}}  & {-9.1{\tiny $\pm$3.3}}   & \bf{62.5{\tiny $\pm$0.9}}  & -3.6{\tiny $\pm$0.9}  & 52.3{\tiny $\pm$1.4} & -5.7{\tiny $\pm$1.3}  \\
O-EWC      & 62.0{\tiny $\pm$0.7}          & -3.2{\tiny $\pm$0.7}  & 54.6{\tiny $\pm$0.7} & -1.3{\tiny $\pm$1.0}   & 54.2{\tiny $\pm$3.1}  & -10.8{\tiny $\pm$3.1}  & \underline{62.4{\tiny $\pm$0.4}}  & {-3.0{\tiny $\pm$0.9}}  & 52.3{\tiny $\pm$1.4} & -5.7{\tiny $\pm$1.3}  \\
HAT        & 63.7{\tiny $\pm$0.7}          & -1.3{\tiny $\pm$0.6}  & 61.4{\tiny $\pm$0.5} & -0.2{\tiny $\pm$0.2}   & 50.1{\tiny $\pm$0.8}  & 0.0{\tiny $\pm$0.1}    & 61.9{\tiny $\pm$1.3}  & -3.2{\tiny $\pm$1.3}  & 61.2{\tiny $\pm$0.7} & -0.1{\tiny $\pm$0.2}  \\
SG-F       & 63.6{\tiny $\pm$1.5}          & 0.0        & 61.5{\tiny $\pm$0.6} & 0.0         & 65.5{\tiny $\pm$1.8}  & 0.0         & 64.1{\tiny $\pm$0.0}  & 0.0        & {62.0{\tiny $\pm$1.3}} & 0.0        \\
MNTDP & 66.3{\tiny $\pm$0.8} & 0.0 & \underline{62.6{\tiny $\pm$0.8}} & 0.0 & 63.1{\tiny $\pm$0.7} & 0.0 &  63.1{\tiny $\pm$0.7} & 0.0 & 63.9{\tiny $\pm$0.5} & 0.0 \\
LMC     & {67.2{\tiny $\pm$1.5}}          & -0.5{\tiny $\pm$0.4}  & {62.2{\tiny $\pm$4.5}} & 2.3{\tiny $\pm$1.6}    & \underline{68.5{\tiny $\pm$1.7}}  & -0.1{\tiny $\pm$0.1}   & 55.1{\tiny $\pm$3.4}  & -7.4{\tiny $\pm$4.0}  & \textbf{63.5{\tiny $\pm$1.9}} & {-1.0{\tiny $\pm$1.5}}  \\
LMC(H)   & 64.9{\tiny $\pm$1.9}          & -0.2{\tiny $\pm$0.2}  & 55.8{\tiny $\pm$2.5} & -0.3{\tiny $\pm$1.2}   & 67.6{\tiny $\pm$2.7}  & -0.8{\tiny $\pm$1.0}   & 54.2{\tiny $\pm$3.6}  & \underline{-2.9{\tiny $\pm$2.0}}  & 53.8{\tiny $\pm$5.7} & 3.1{\tiny $\pm$5.5}   \\
ER      & 62.9{\tiny $\pm$0.4}          & -0.7{\tiny $\pm$1.1}  & 55.9{\tiny $\pm$1.2} & 1.7{\tiny $\pm$0.9}    & 54.8{\tiny $\pm$3.2}  & -4.2{\tiny $\pm$3.7}   & 60.7{\tiny $\pm$1.0}  & -1.5{\tiny $\pm$0.5}  & 55.6{\tiny $\pm$1.3} & -1.2{\tiny $\pm$1.5}  \\  
DER++      & 65.3{\tiny $\pm$0.5}          & 0.2{\tiny $\pm$0.7}  & 60.2{\tiny $\pm$1.4} & 0.9{\tiny $\pm$0.5}    & 59.0{\tiny $\pm$3.8}  & -3.8{\tiny $\pm$2.0}   & 64.7{\tiny $\pm$1.4}  & 0.4{\tiny $\pm$0.2}  & 58.5{\tiny $\pm$1.2} & 0.0{\tiny $\pm$1.7}  \\  
CLSER      & 63.9{\tiny $\pm$0.7}          & -0.1{\tiny $\pm$0.9}  & 56.7{\tiny $\pm$1.1} & 1.9{\tiny $\pm$0.7}    & 60.2{\tiny $\pm$1.8}  & -3.6{\tiny $\pm$1.7}   & 61.3{\tiny $\pm$1.7}  & -0.2{\tiny $\pm$0.9}  & 58.2{\tiny $\pm$1.1} & -0.4{\tiny $\pm$1.1}  \\
Cassle      & 66.4{\tiny $\pm$1.2}          & -3.2{\tiny $\pm$1.0}  & 62.1{\tiny $\pm$0.7} & 2.5{\tiny $\pm$0.5}    & 61.7{\tiny $\pm$1.2}  & -5.7{\tiny $\pm$0.8}   & 65.8{\tiny $\pm$0.9}  & -3.5{\tiny $\pm$1.5}  & 58.8{\tiny $\pm$0.5} & -1.2{\tiny $\pm$0.5}  \\
\midrule
DualNet & \underline{68.1{\tiny $\pm$0.7}}          & \underline{0.2{\tiny $\pm$0.4}}   & {62.2{\tiny $\pm$1.0}} & \textbf{4.6{\tiny $\pm$1.3}}    & 63.8{\tiny $\pm$1.6}  & \underline{-3.4{\tiny $\pm$0.8}}   & \underline{67.3{\tiny $\pm$0.7}}  & \textbf{0.7{\tiny $\pm$1.5}}  & 59.6{\tiny $\pm$0.9} & {-1.2{\tiny $\pm$0.1}}  \\ 
DualNet++ & \textbf{69.6{\tiny $\pm$1.1}}          & \textbf{1.3{\tiny $\pm$0.8}}   & \textbf{63.3{\tiny $\pm$0.9}} & \underline{4.3{\tiny $\pm$0.5}}    & {64.1{\tiny $\pm$2.2}}  & \textbf{-3.4{\tiny $\pm$1.1}}   & \textbf{68.1{\tiny $\pm$0.5}}  & \textbf{0.7{\tiny $\pm$1.8}}  & \underline{62.2{\tiny $\pm$0.6}} & \textbf{0.1{\tiny $\pm$1.1}}  \\ \bottomrule
\end{tabular}}
\end{table*}

\subsubsection{Evaluation Metrics on CTrL}
Tab.~\ref{tab:ctrl-result} reports the evaluation metrics at the end of training on the CTrL benchmarks. \blue{We organize the results into three blocks: (i) task-free setting with a randomly initialized backbone; (ii) task-free setting with a pre-trained network from ImageNet; and (iii) task-aware setting with a randomly initialized network.}
In general, the results show that dynamic architecture methods, especially recent works such as the shared-head variants of MNTDP~\cite{veniat2020efficient} and LMC~\cite{ostapenko2021continual}, can perform competitively on this benchmark and outperforms the static architecture approaches. 
\blue{Among replay-based methods, we observe that, compared to the vanilla ER, CLSER effectively handled different distribution shifts while offering limited improvements in handling limited training samples for learning from unrelated tasks. \red{On the other hand, Cassle can slightly outperforms ER and DER++ in some cases thanks to the SSL learning component, which is optimized for many iterations in this batch learning setting. However, both CLSER and Cassle still perform worse than our DualNets, which shows the benefits of our fast-and-slow learning framework in addressing challenging transfer scenarios.} Overall, DualNet and DualNet++ outperform all task-free baselines considered with only one exception in the \Sout stream where they were outperformed by EWC and O-EWC. Moreover, in most cases, DualNet++ achieves improvements on \ACC and \BWT over DualNet, indicating the benefits of its spatial dropout layers in preventing negative transfer in continual learning. In the task-aware setting, we observe many competitive baselines with high performance. Notably, most baselines in this category are dynamic architecture approaches, which have long dominated the CTrL benchmark. 
Nevertheless, DualNet++ can perform favorably and achieve top-2 performances in several streams. To the best of our knowledge, this is the first result showing a static architecture method consistently performing comparable with the dynamic architecture ones on the CTrL benchmarks. 
We also highlight that although they achieved strong results, dynamic architecture methods in the Task-aware category exhibit high variance on different runs, which arises from them training larger models on a small number of samples. On the other hand, DualNets perform consistently and have small variances in all cases.}

\blue{We also select ER and DER++ and compare them with DualNets when using the pre-trained ImageNet as an initialization. We observe that this strategy generally improves the performances in scenarios where the first task has limited training data (\Splus), there are input distribution shifts (\Sin), or learning from unrelated tasks (\Spl). On the other hand, when the first task has abundant training samples (\Sminus), using a pre-trained network only offered marginal improvements. Lastly, the \Sout~stream has sufficient training samples in the first task and introduces a label permutation, which a pre-trained network may not address. In this case, we did not observe significant differences between using a pre-trained network or training from scratch.}

\begin{table*}[t]
\centering
\caption{\red{Transfer results on the CTrL benchmark. All methods are task-aware. We report the accuracy of the first and last tasks of each stream and $\Delta$, the difference between the last task accuracy from the model compared to the reference \textbf{Independent} model. We highlight the best mean metrics in bold, and underline the second best results}}
\label{tab:transfer}
\begin{tabular}{lcccccccccccc}
\toprule
\multirow{2}{*}{Method}      & \multicolumn{3}{c}{$\mc S^-$}            & \multicolumn{3}{c}{$\mc S^+$}    & \multicolumn{3}{c}{$\mc S^{\mathrm{in}}$}    & \multicolumn{3}{c}{$\mc S^{\mathrm{out}}$}         \\ \cmidrule{2-13}
   & ACC T1     & ACC T6     & $\Delta$ & ACC T1     & ACC T6     & $\Delta$ & ACC T1     & ACC T6     & $\Delta$ & ACC T1     & ACC T6      & $\Delta$ \\ \midrule
Independent  & 65.5$\pm$0.7 & 41.8$\pm$1.0 & 0     & 41.3$\pm$2.9 & \underline{65.6$\pm$0.5} & 0     & 98.5$\pm$0.2 & 76.9$\pm$4.9 & 0     & 65.9$\pm$0.6 & 43.5$\pm$1.6  & 0     \\
\red{MNTDP}  & 63.0$\pm$3.6 & 56.9$\pm$5.1 & 15.1    & \textbf{43.2$\pm$0.7} & \textbf{65.9$\pm$0.8} & \textbf{0.3}   & \textbf{98.9$\pm$0.1} & \underline{93.3$\pm$1.6} & {16.4}  & 65.0$\pm$1.2 & 57.7$\pm$1.7  & 14.2  \\
\red{LMC}     & 65.2$\pm$0.4 & 60.0$\pm$1.1 & 18.2  & 42.9$\pm$0.9 & 60.6$\pm$1.9 & -4.7  & \underline{98.7$\pm$0.1} & 92.5$\pm$7.6 & 15.6  & 65.2$\pm$0.2 & 59.8$\pm$1.1  & 16.3  \\
\red{LMC(H)}  & 62.2$\pm$0.4 & 63.0$\pm$1.7 & 21.2  & \underline{43.1$\pm$0.6} & 62.2$\pm$0.7 & -3.4  & \underline{98.7$\pm$0.1} & 88.3$\pm$1.6 & 11.4  & 65.5$\pm$0.6 & 42.0$\pm$21.9 & -1.5  \\
\red{SG-F} & 64.9$\pm$0.4 & 49.1$\pm$7.3 & 7.3   & \underline{43.1$\pm$0.4} & 61.7$\pm$1.7 & -3.9  & 98.8$\pm$0.1 & 80.4$\pm$6.8 & 3.5   & 65.0$\pm$0.4 & 51.5$\pm$6.5   & 8     \\
\red{DER++} & 68.5$\pm$1.5 & 69.9$\pm$0.8 & 28.1   & {36.7$\pm$2.4} & 58.8$\pm$1.7 & -6.8  & 98.1$\pm$0.2 & 93.4$\pm$2.3 & 16.5   & 67.4$\pm$2.3 & 66.9$\pm$2.0   & 23.4     \\
\midrule
\red{DualNet} & \underline{71.8$\pm$0.8} & \underline{71.9$\pm$0.8} & \textbf{27.2}  & 41.2$\pm$0.4 & 64.5$\pm$0.8 & {-1.1}  & \textbf{98.9$\pm$0.1} & \textbf{94.8$\pm$2.4} & {\bf 17.9}   & \underline{69.6$\pm$1.8} & \underline{68.4$\pm$1.7}  & \underline{24.9} \\
\red{DualNet++} & \textbf{72.6$\pm$0.9} & \textbf{72.8$\pm$0.6} & \textbf{31.0}  & 40.0$\pm$0.1 & 64.8$\pm$0.8 & {-0.8}  & \textbf{98.9$\pm$0.1} & \textbf{94.8$\pm$1.5} & \textbf{17.9}   & \textbf{71.0$\pm$1.4} & \textbf{71.4$\pm$1.2}  & \textbf{27.9} \\\bottomrule
\end{tabular}
\end{table*}

\begin{table*}[ht]
\centering
\caption{DualNet++ with different dropout ratio $p$ on the CTrL benchmark using the task-aware evaluation. With the ratio $p=0.0$, DualNet++ reduces to the standard DualNet. Best mean results are highlighted in bold} \label{tab:drop}
\begin{tabular}{lcccccccccc}
\toprule
\multirow{2}{*}{DualNet++(A)} & \multicolumn{2}{c}{\Sminus}  & \multicolumn{2}{c}{\Splus} & \multicolumn{2}{c}{\Sin} & \multicolumn{2}{c}{\Sout} & \multicolumn{2}{c}{\Spl} \\ \cmidrule{2-11}
                      & \ACC        & \BWT        & \ACC        & \BWT       & \ACC        & \BWT        & \ACC         & \BWT        & \ACC         & \BWT       \\ \midrule
$p=0.0$                     & 68.1{\tiny $\pm$0.7} & 0.2{\tiny $\pm$0.4}  & 62.2{\tiny $\pm$1.0} & 4.6{\tiny $\pm$1.3} & 63.8{\tiny $\pm$1.6} & -3.4{\tiny $\pm$0.8} & 67.3{\tiny $\pm$0.7}  & 0.7{\tiny $\pm$1.5}  & 58.6{\tiny $\pm$0.9}  & -1.2{\tiny $\pm$0.1} \\
$p=0.1$                    & 68.8{\tiny $\pm$0.8} & 0.7{\tiny $\pm$0.3}  & 61.6{\tiny $\pm$1.1} & 1.3{\tiny $\pm$0.7} & 63.6{\tiny $\pm$2.0} & \textbf{4.7{\tiny $\pm$0.6}}  & \textbf{69.1{\tiny $\pm$1.4}}  & \textbf{1.4{\tiny $\pm$0.6}}  & 61.5{\tiny $\pm$1.8}  & 1.4{\tiny $\pm$0.6} \\
$p=0.2$                    & \textbf{69.6{\tiny $\pm$1.1}} & \textbf{1.3{\tiny $\pm$0.8}}  & \textbf{63.3{\tiny $\pm$0.9}} & \textbf{4.3{\tiny $\pm$0.5}} & \textbf{64.1{\tiny $\pm$2.2}} & -3.4{\tiny $\pm$1.1} & 68.1{\tiny $\pm$0.5}  & 0.7{\tiny $\pm$1.8}  & \textbf{62.2{\tiny $\pm$0.6}}  & 0.1{\tiny $\pm$1.1}\\
$p=0.3$                    & 66.8{\tiny $\pm$1.2} & -1.0{\tiny $\pm$0.7} & 61.2{\tiny $\pm$0.6} & 3.5{\tiny $\pm$0.2} & 63.5{\tiny $\pm$1.6} & -2.6{\tiny $\pm$0.7} & 67.6{\tiny $\pm$0.8}  & 0.4{\tiny $\pm$0.5}  & 61.3{\tiny $\pm$1.4}  & 0.6{\tiny $\pm$0.5} \\
$p=0.5$                    & 67.1{\tiny $\pm$0.6} & -0.7{\tiny $\pm$1.1} & 60.7{\tiny $\pm$1.0} & 3.6{\tiny $\pm$0.3} & 63.2{\tiny $\pm$0.2} & -1.8{\tiny $\pm$1.1} & 67.2{\tiny $\pm$1.2}  & -0.2{\tiny $\pm$0.7} & 61.2{\tiny $\pm$0.9}  & 0.7{\tiny $\pm$0.1} \\ \bottomrule
\end{tabular}
\vspace*{-0.1in}
\end{table*}

\subsubsection{Transfer Results on CTrL}
We now take a closer look at the transferring capabilities of different methods on the CTrL benchmark. Recall that in the \Sminus, \Splus, \Sin, \Sout~streams, only the first and last tasks are closely related, while the intermediate ones are distractors. Therefore, the \ACC metric alone might not be sufficient to evaluate the model's ability to transfer because it also measures the performance of unrelated tasks. Thus, in these streams, it is helpful to look at the accuracy of the first and last task explicitly~\cite{ostapenko2021continual}, and compare them with a reference model, Independent, that trains a separate model for each task. We report the results of this experiment in Tab.~\ref{tab:transfer}. 
We can see that except for the \Splus stream, both DualNet and DualNet++ achieve significantly better accuracy on the last task and have a higher differences ($\Delta$) compared to the reference model. On the \Splus stream, we observe that DualNets are marginally worse than a few baselines, suggesting that remembering and continuing to learn from a limited representation is challenging for DualNets. This phenomenon is easy to understand since learning good representations from limited data with distractors is a challenging problem. Overall, the results suggest that DualNets can remember long-term knowledge in several cases and learn robust representations to distribution shifts, which facilitates successful continual learning in complex scenarios. Moreover, retaining and continuing learning from limited experiences (the \Splus stream) presents a promising future research direction.

\subsubsection{Robustness to The Dropout Ratio}

In literature, the dropout ratio for fully connected layers are commonly set as $p=0.5$ while this value is set to lower values, e.g. $p=0.1$ or $p=0.15$, for convolutional layers~\cite{tompson2015efficient}. We now investigate how this hyper-parameter affects DualNet++ performance. We consider the task-aware setting and report DualNet++ with different dropout ratios in Tab.~\ref{tab:drop}.
It is worth noting that by removing the dropout layer (setting $p=0.0$), DualNet++ reduces to DualNet. The results show that DualNet++ is robust to and beneficial from the small dropout ratios. Specifically, DualNet++ achieves similar performances when $p=0.1$ and $p=0.2$, and both can outperform the standard DualNet ($p=0.0$). 
When using larger dropout ratio, e.g. $p=0.3$ and $p=0.5$, we observe a performance drop on all streams in the CTrL benchmark, which is consistent with the conventional usage of dropout layers in convolutional networks.

\subsection{Summary of Results}
We now provide a summary of the experimental results.

We have conducted experiments on various continual learning settings and examined different scenarios, ranging from the traditional settings to the complex scenarios of semi-supervised learning or complex transfer scenarios. In most cases, DualNet and DualNet++ outperform the baselines, often quite significantly. There is an exception of the \Splus and \Sin streams where the model needs to learn from limited data with different input distributions, which presents an interesting future work. We also found DualNets to be robust to the choice of SSL loss, an benefited from the LA optimizer and more SSL training iterations in the online continual learning setting. Between DualNet and DualNet++, DualNet++ achieves similar performances to DualNet in the controlled environment where labeled data is plentiful. On the other hand, DualNet++ offers significant improvements in scenarios where labeled data is limited (semi-supervised setting) or there exists negative knowledge transfer from unrelated tasks (CTrL benchmark). 

\vspace*{-0.05in}
\section{Conclusion}\label{sec:conclusion}
Inspired by the Complementary Learning System theory, we propose a novel fast-and-slow learning framework for continual learning and conceptualize it into the DualNets paradigm. DualNets (DualNet and DualNet++) comprise two key learning components: (i) a slow learner that focuses on learning a general and task-agnostic representation using the memory data; and (ii) a fast learner focuses on capturing new supervised learning knowledge via a novel adaptation mechanism. Moreover, the fast and slow learners complement each other while working synchronously, resulting in a holistic continual learning method. Our experiments on challenging benchmarks demonstrate the efficacy of DualNets. Lastly, extensive and carefully designed ablation studies show that DualNets are robust the hyper-parameter configurations, scalable with more resources, and can work well in several challenging continual learning scenarios.

\textbf{Limitations and Future Work } Because the DualNet's slow learner can always be trained in the background, it incurs computational costs that need to be properly managed. For large-scale systems, such additional computations can increase infrastructural costs substantially. Therefore, it is important to manage the slow learner to balance between performance and computational overheads. In addition, implementing DualNet to specific applications requires additional considerations to address its inherent challenges. For example, medical image analysis applications may require paying attention to specific regions in the image or considering the data imbalance. However, since we demonstrate the efficacy of DualNet in general settings, such properties are not considered. In practice, it would be more beneficial to capture such domain-specific information to achieve better results. For general applications, \red{ we also expect that a more sophisticated objective to train the slow learner can further improve the results or greatly reduces the training complexity, which is an important direction towards reducing the training costs of DualNets.} 
Lastly, through extensive experiments, we identified the scenario of \Sin and \Splus, continual learning from limited data under distribution shifts, to be challenging for DualNets. This suggests a promising future research direction to develop a better, more robust representation learning from limited training samples and can be robust to distribution shifts.


%

\vspace*{-0.05in}
\appendices
\begin{algorithm*}[!ht]
	\DontPrintSemicolon
	\SetKwFunction{algo}{TrainDualNet}
	\SetKwFunction{proc}{Look-ahead}\SetKwFunction{procc}{MemoryUpdate}
	\SetKwProg{myalg}{Algorithm}{}{}
	\myalg{\algo{$\bm \theta, \bm \phi, \mc D^{tr}_{1:T}$}}{
		\kwRequire {slow learner $\bm \phi$, fast learner  $\bm \theta$, episodic memory $\mc M$, inner updates $N$, Look-ahead inner updates $K$ (for Look-ahead)}
		\kwInit{$\bm \theta, \bm \phi, \mc M \gets \varnothing$}
		\For{$t \gets 1$ \textbf{to} $T$ }{
			\For(\tcp*[f]{Receive the dataset $\Dtrain{t}$ sequentially}){$j \gets 1$ \textbf{to} $n_{batches}$} { 
				{Receive a mini batch of data $\mc B_j$ from $\mc D^{tr}_t$ }\;
				{$\mc M \leftarrow \textrm{\bf MemoryUpdate}(\mc M, \mc B_j)$} \tcp*[f]{Update the episodic memory}
				
    			\For(\tcp*[f]{Train the slow learner \textcolor{red}{synchronously}}){$i \gets 1$ \textbf{to} $\infty$} {
    				{Train the slow learner using the \texttt{Look-ahead procedure}}
    			}
    			\For(\tcp*[f]{Train the fast learner \textcolor{red}{synchronously}}){$n \gets 1$ \textbf{to} $N$} {
    				{$\bm M_n \leftarrow$ Sample($\mc M$)}\;
    				{$\mc B_n \leftarrow \bm M_n \cup \mc B_j$}\;
    				{SGD update the slow learner: $\bm \phi \leftarrow \bm \phi - \nabla_{\bm \phi} \mc L_{tr}(\mc B_n) $} \;
    				{SGD update the fast learner $\bm \theta \leftarrow \bm \theta - \nabla_{\bm \theta} \mc L_{tr}(\mc B_n) $}\;
			    }
            }
			{$\mc M^{em}_t \leftarrow \mc M^{em}_t \cup \{\pi(\hat{y}/\tau) \}$}\;
		}
		\Return $\bm \theta, \bm \phi$}
		
	\setcounter{AlgoLine}{0}
	\SetKwProg{myproc}{Procedure}{}{}
	\myproc{\proc{$\bm \phi, \mc M$}}{
	    {$\tilde{\bm \phi_0} \gets \bm \phi$}\;
		\For{$k \gets 1$ \textbf{to} $K-1$}{
            {$\bm M_k \leftarrow$ Sample($\mc M$) $\cup \mc B_j$}\;
		    {Obtains two views of $\bm M_k: \bm M^A_k, \bm M^B_k$}\;
		    {Calculate the Barlow Twins loss: $\mc L_{\mc B \mc T}(\tilde{\phi}_k,\bm M^A_k, \bm M^B_k)$}\;
		    {SGD update the slow learner: $\tilde{\bm \phi}_{k+1} \gets \bm \phi_{k} - \epsilon \nabla_{\bm \phi_{k}} \mc L_{\mc B \mc T} $}
		}
		{Look-ahead update the slow learner: $\bm \phi \gets \bm \phi + \beta (\tilde{\bm \phi}_K - \bm \phi)$}\;
		\Return {$\bm \phi$} 
	}
	\caption{Pseudo-code to train DualNets.}
	\label{alg:pseudo}
\end{algorithm*}

\section{DualNets Pseudo code} \label{app:pseudo}
We provide DualNets' pseudo-code in Algorithm~\ref{alg:pseudo}.

\vspace*{-0.05in}
\ifCLASSOPTIONcompsoc
  \section*{Acknowledgments}
\else
  \section*{Acknowledgment}
\fi

The first author, Quang Pham, gratefully acknowledges the support by the Lee Kuan Yew Fellowship awarded by Singapore Management University.

\ifCLASSOPTIONcaptionsoff
  \newpage
\fi

\bibliographystyle{IEEEtran}
\bibliography{IEEEabrv,ref}

%

\begin{IEEEbiography}[{\includegraphics[width=1in,height=1.25in,clip,keepaspectratio]{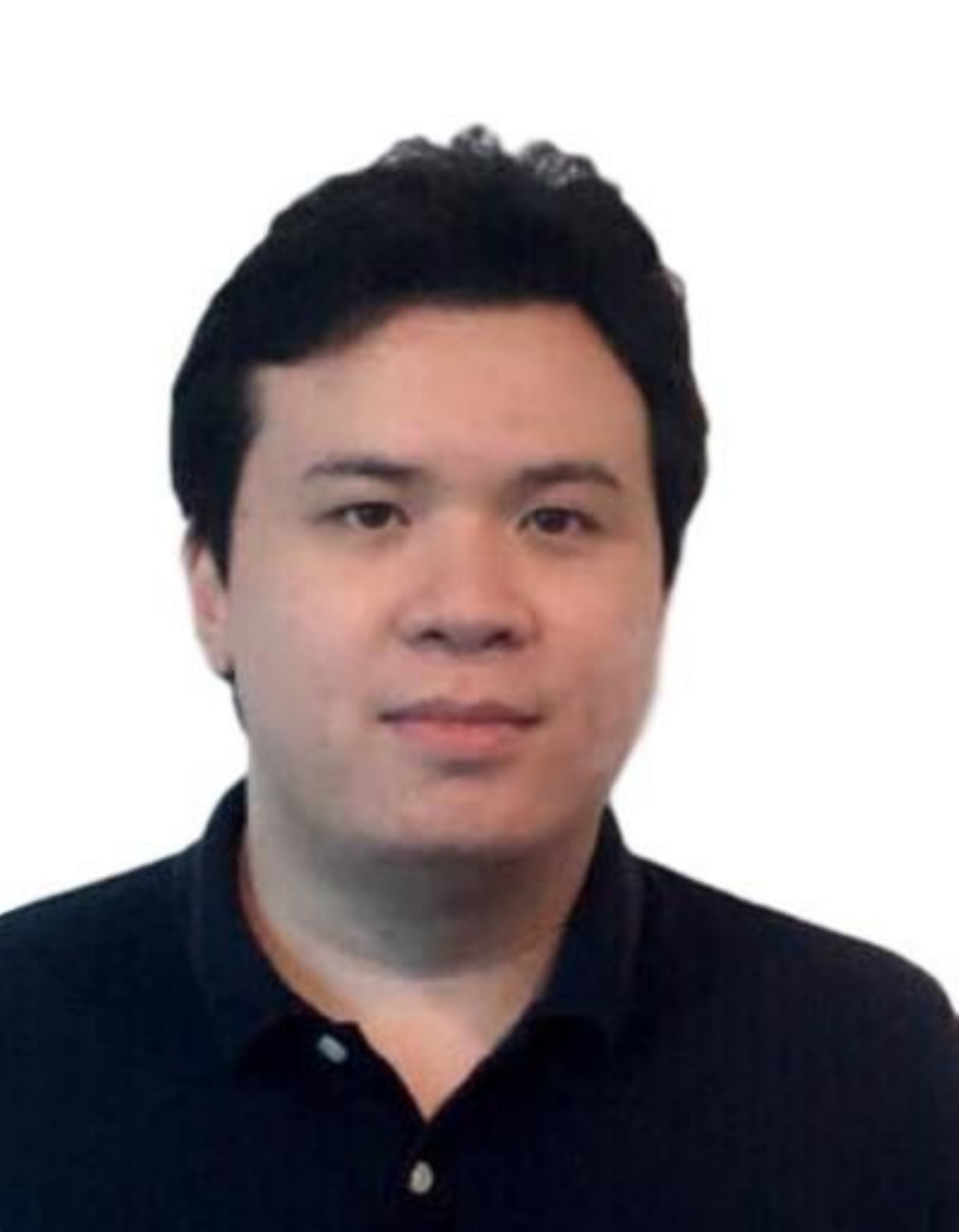}}]{Quang Pham}
Quang Pham is currently a research scientist at Institute for Infocomm Research (I$^2$R), Agency for Science, Technology and Research (A*STAR). He obtained his PhD degree from School of Computing and Information Systems at Singapore Management University. His research interests include continual learning and deep learning.
\end{IEEEbiography}

\begin{IEEEbiography}[{\includegraphics[width=1in,height=1.25in,clip,keepaspectratio]{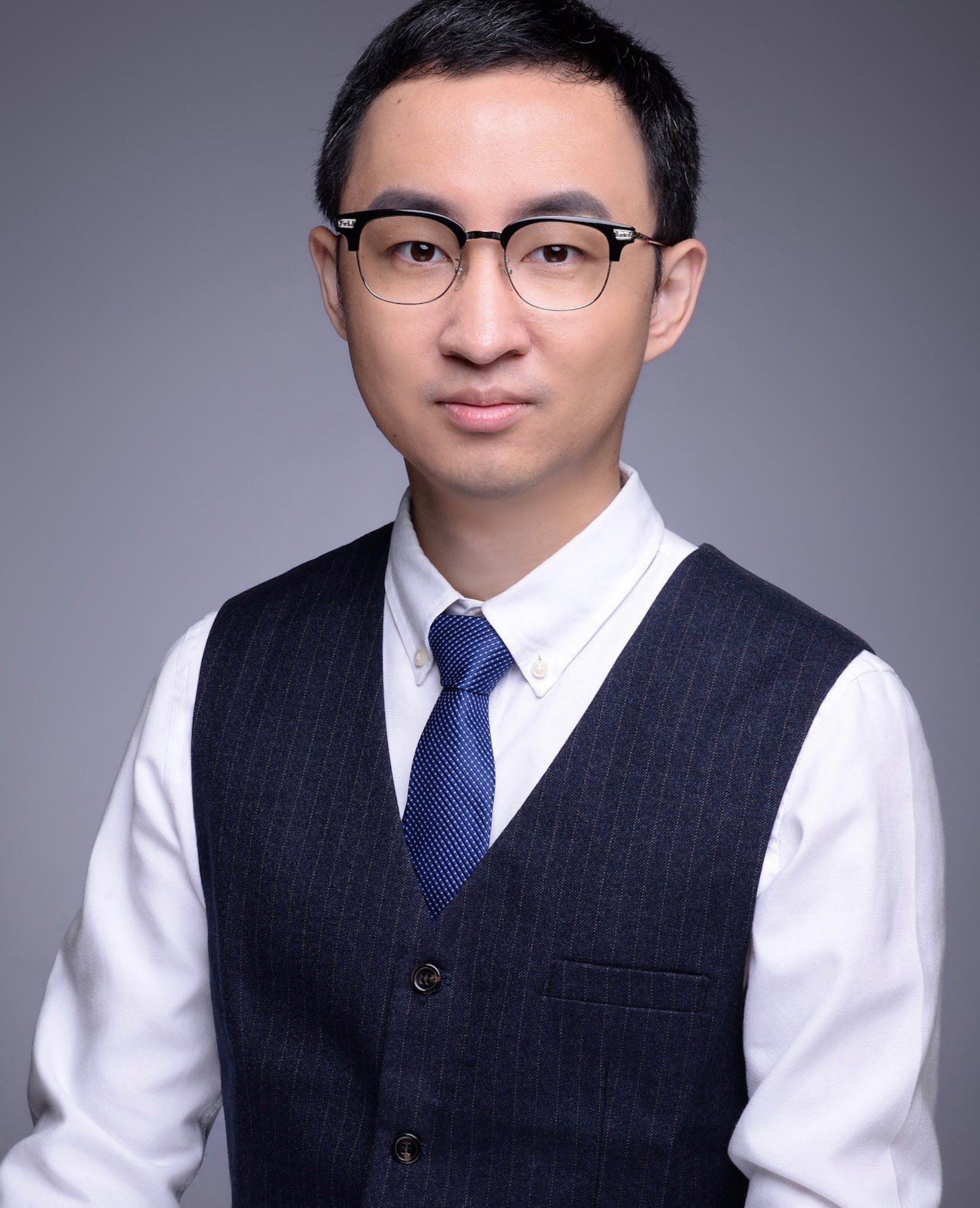}}]{Chenghao Liu}
Chenghao Liu is currently a senior applied scientist of Salesforce Research Asia. Before, he was a research scientist in the School of Information Systems (SIS), Singapore Management University (SMU), Singapore. He received his Bachelor degree and Ph.D degrees from the Zhejiang University. His research interests include large-scale machine learning (online learning and deep learning) with application to tackle big data analytics challenges across a wide range of real-world applications.
\end{IEEEbiography}


\begin{IEEEbiography}[{\includegraphics[width=1in,height=1.25in,clip,keepaspectratio]{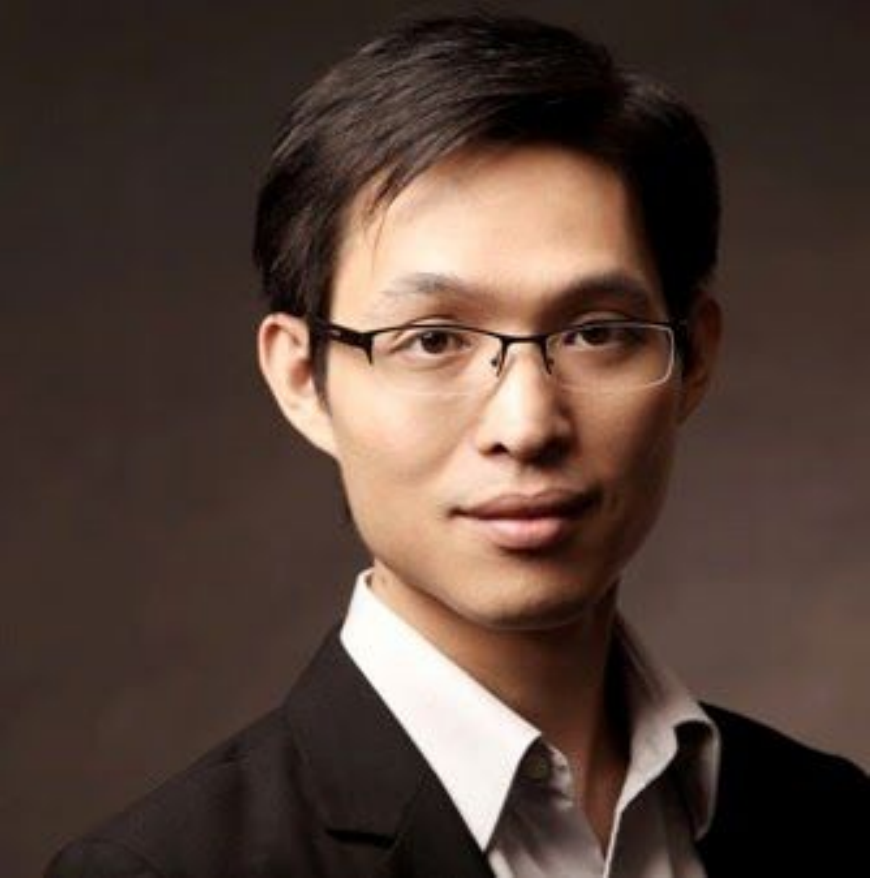}}]{Steven C. H. Hoi}
Steven C. H. Hoi is currently the Managing Director of Salesforce Research Asia, and a Professor of Information Systems at Singapore Management University, Singapore. He received his Bachelor degree from Tsinghua University, P.R. China, in 2002, and his Ph.D degree in computer science and engineering from The Chinese University of Hong Kong, in 2006. He has served as the Editor-in-Chief for Neurocomputing Journal, general co-chair for ACM SIGMM Workshops on Social Media, program co-chair for the fourth Asian Conference on Machine Learning, book editor for “Social Media Modeling and Computing”, guest editor for ACM Transactions on Intelligent Systems and Technology. He is an IEEE Fellow and ACM Distinguished Member.
\end{IEEEbiography}
\vfill



\end{document}